\documentclass{article}
\usepackage[preprint]{neurips_2026}
\usepackage[utf8]{inputenc}
\usepackage[T1]{fontenc}
\usepackage{courier}
\usepackage{hyperref}
\hypersetup{hidelinks}
\usepackage{url}
\usepackage{booktabs}
\usepackage{amsfonts}
\usepackage{nicefrac}
\usepackage{microtype}
\usepackage[table]{xcolor}
\usepackage{amsmath}
\usepackage{amssymb}
\usepackage{mathtools}
\usepackage{amsthm}
\usepackage{multirow}
\usepackage{graphicx}
\usepackage{subcaption}
\usepackage{caption}
\usepackage{algorithm}
\usepackage{algorithmic}
\usepackage{placeins}
\usepackage{wrapfig}
\newcommand{\Needspace}[1]{%
  \par\begingroup
  \dimen0=#1\relax
  \dimen2=\pagegoal\relax
  \advance\dimen2 by -\pagetotal\relax
  \ifdim\dimen0>\dimen2\relax\newpage\fi
  \endgroup
}
\newtheorem{proposition}{Proposition}
\title{LEAP: Unlocking dLLM Parallelism via Lookahead Early-Convergence Token Detection}

\author{
  Haohui Zhang \\
  Shanghai Jiao Tong University\\
  \texttt{zhanghaohui@sjtu.edu.cn} \\
  \And
  Zhiye Wang \\
  Shanghai Jiao Tong University\\
  \texttt{21-wzy@sjtu.edu.cn} \\
  \And
  Xiaoying Gan \\
  Shanghai Jiao Tong University\\
  \texttt{ganxiaoying@sjtu.edu.cn} \\
  \And
  Xinbing Wang \\
  Shanghai Jiao Tong University\\
  \texttt{xwang8@sjtu.edu.cn} \\
  \And
  Bo Jiang\thanks{Corresponding author} \\
  Shanghai Jiao Tong University\\
  \texttt{bjiang@sjtu.edu.cn} \\
}

\begin{document}
\maketitle
\raggedbottom

\begin{abstract}
  Diffusion Language Models (dLLMs) enable highly parallel generation. Existing parallel decoders select high-confidence marginal predictions to limit errors from committing dependent tokens together. However, the stringent confidence thresholds required to preserve accuracy severely constrain the scalability of parallelism. Token-level analysis reveals that many predictions match their eventual confidence-based decoding outputs early, before meeting standard confidence thresholds, suggesting that these criteria are conservative. In response, we introduce LEAP (Lookahead Early-Convergence Token Detection for Accelerated Parallel Decoding). LEAP is a training-free, plug-and-play method that leverages future context filtering and multi-sequence superposition to detect early-converging tokens. Tokens that remain invariant under the superposed context are decoded early, with a loose confidence threshold filtering high-entropy predictions. Benchmarking across diverse domains shows that LEAP lowers inference latency and decoding steps. Compared to confidence-based decoding, the average number of denoising steps is reduced by about 30\%. On the GSM8K dataset, combining LEAP with dParallel accelerates decoding to 7.2 tokens per step while preserving model precision. LEAP effectively breaks the reliance on high-confidence priors, offering a novel paradigm for parallel decoding.
\end{abstract}

\begin{figure}[!t]
  \centering
  \includegraphics[width=0.66\textwidth]{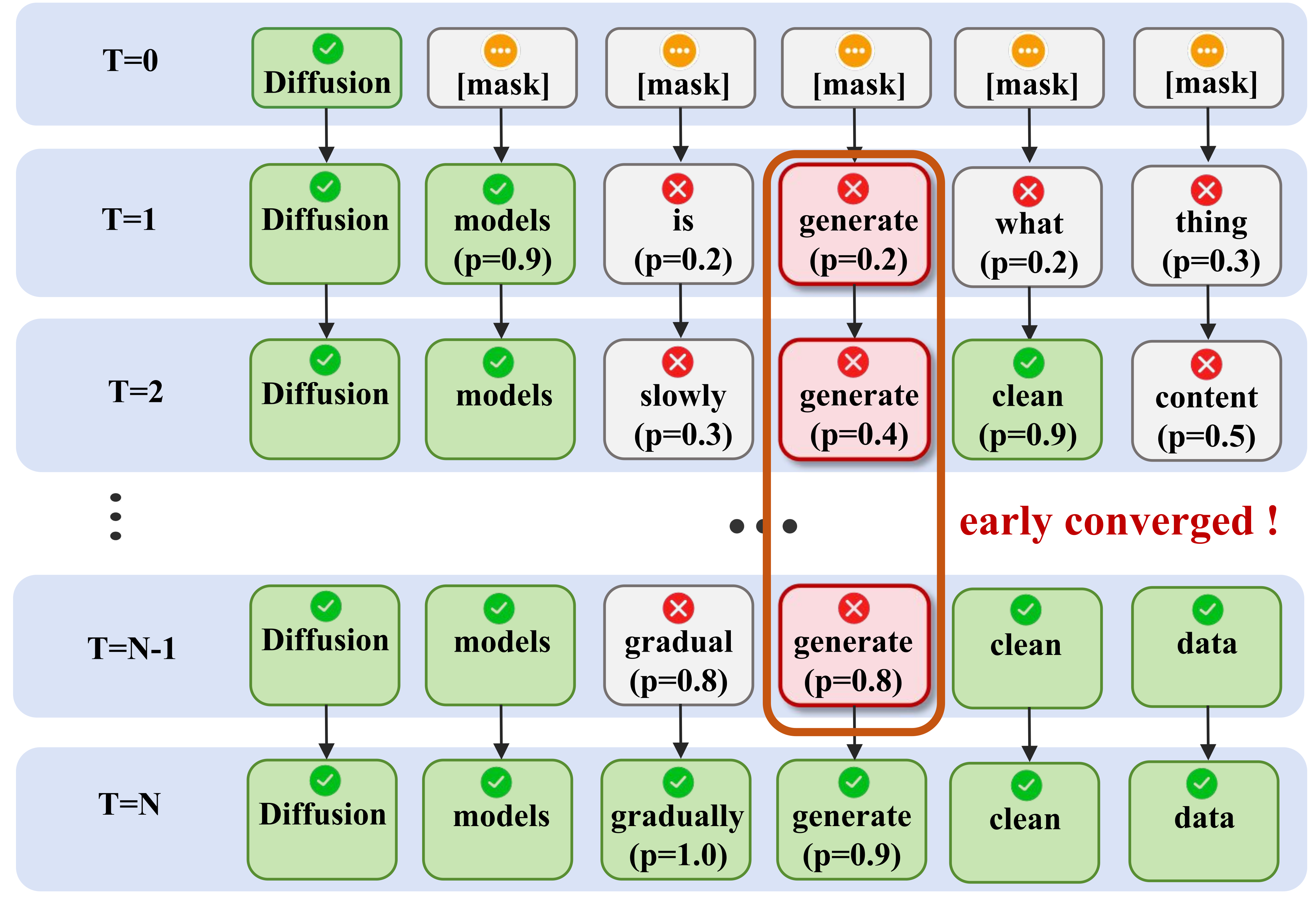}
  \caption{Early convergence in diffusion language models. Across denoising steps $T=0$ to $T=N$, confidence-based decoding commits only high-confidence tokens (green). The token ``generate'' matches the final reference and is stable from $T=1$ but remains masked until late because its confidence stays below the threshold.}
  \label{motivation_example}
\end{figure}

\section{Introduction}
Autoregressive large language models (AR-LLMs) have long dominated the field of language modeling  \citep{achiam2023gpt, yang2025qwen3, liu2024deepseek}. However, their inherent sequential generation process restricts further improvements in inference speed. Consequently, diffusion large language models (dLLMs) have emerged as a promising new paradigm   \citep{nie2025large, bie2025llada2, ye2025dream, liu2025wedlm}, garnering significant attention for their potential to highly parallel generation. Recent diffusion and diffusion--autoregressive systems achieve inference speeds exceeding autoregressive baselines by reducing per-step costs and increasing parallel generation \citep{wang2025diffusion, liu2025tidar, liu2025wedlm}.

Existing parallel decoders often select multiple high-confidence tokens from marginal predictions \citep{wu2025fast}. Committing these predictions together can lose dependencies that would otherwise be resolved by subsequent context updates, reducing quality on tasks requiring strict logical coherence. Conservative selection preserves quality but restricts the number of tokens decoded per step. 

For greedy decoding, Fast-dLLM establishes a sufficient high-confidence condition under which jointly selected marginal modes agree with the joint mode \citep{wu2025fast}. In practice, few tokens meet conservative thresholds, while simply relaxing the threshold can reduce task accuracy. 

We identify two limitations of relying on confidence alone. First, confidence-based decoding limits parallel candidates to high-confidence tokens, yet many medium-confidence predictions already match their eventual CBPD outputs. High thresholds can therefore delay stable predictions. Second, a high-probability prediction has low surprisal, $-\log p$, motivating information-based analyses of parallel decoding \citep{fu2025bits}. The practical challenge is to expand parallelism and provide useful context earlier without sacrificing accuracy.

We introduce \textbf{L}ookahead \textbf{E}arly-Convergence Token Detection for \textbf{A}ccelerated \textbf{P}arallel Decoding (LEAP), a training-free, plug-and-play parallel decoding strategy. We find empirically that many medium-confidence predictions match the CBPD reference early and remain stable, which implies that a large number of forward steps are performing repeated predictions on them. These tokens show low sensitivity—their predictions stabilize early and demonstrate robustness to future contextual changes. Capitalizing on this, we propose a convergence detection strategy based on future context perturbation. By contrasting the predictions of the current context with those of a superposed context containing potential future information, we identify tokens that exhibit low sensitivity and high robustness to future updates, thereby enabling their early decoding. The feasibility of this strategy stems from the novel future context candidate pruning and multi-sequence superimposed consistency detection strategy we propose. 
Decoding stable medium-confidence tokens expands the current step's parallelism and makes their predictions available as context earlier, allowing subsequent steps to condition on a more complete sequence.

We conduct extensive evaluations on two popular open-source dLLMs, LLaDA and Dream, covering mathematics, code generation, and multi-disciplinary QA. Empirical results show that LEAP improves generation parallelism across all benchmarks, decreasing latency by around 24\% against confidence-based decoding strategy. Meanwhile, LEAP slightly improves average accuracy on LLaDA. Further analysis confirms that LEAP establishes a better Pareto frontier between speed and accuracy compared to confidence-based decoding strategy.

\section{Related Work}
\textbf{Diffusion Language Models.} Diffusion large language models (dLLMs) replace autoregressive factorization with iterative denoising. LLaDA \citep{nie2025large} and LLaDA 2.0 \citep{bie2025llada2} show that masked diffusion models can match 8B autoregressive baselines such as LLaMA 3 \citep{dubey2024llama} and scale to a 100B-parameter MoE. Dream \citep{ye2025dream} improves downstream accuracy while retaining parallel generation by initializing from autoregressive weights. Industry-developed systems and model previews, including Gemini Diffusion \citep{deepmind2025geminidiffusion}, Mercury \citep{inception2025mercury}, and Seed Diffusion \citep{song2025seed}, further demonstrate the latency benefits of diffusion decoding.

\textbf{Acceleration of dLLMs.} Existing methods either reduce per-step cost or decode more tokens per step. Fast-dLLM-Cache \citep{wu2025fast} and dKV-Cache \citep{ma2025dkv} approximate caching for bidirectional attention, whereas ReFusion \citep{li2025refusion} combines diffusion-based slot selection with autoregressive infilling, and WeDLM \citep{liu2025wedlm} supports diffusion decoding with standard causal attention. To increase parallelism, Fast-dLLM-Parallel \citep{wu2025fast} and EB-Sampler \citep{ben2025accelerated} select tokens by confidence and entropy. D2F \citep{wang2025diffusion} and dParallel \citep{chen2025dparallel} use distillation; Prophet \citep{li2025diffusion} uses confidence gaps for early commitment; KLASS \citep{kim2026klass} combines confidence with inter-step KL divergence; and LoPA \citep{xu2025lopa} selects among sampled future branches. These approaches still favor high-confidence or low-entropy tokens, or require multi-branch search. LEAP instead detects medium-confidence tokens that remain stable under future-context perturbations, without training.

\section{Methodology}

\Needspace{5\baselineskip}
\subsection{Preliminary}
\subsubsection{Diffusion Language Models}
Diffusion Language Models (DLMs) model text generation as a discrete diffusion process involving a forward corruption phase and a reverse denoising phase. The forward process corrupts a clean sequence $\mathbf{x}_0$ into a masked state $\mathbf{x}_t$ at timestep $t$. This can be formulated as a marginal distribution where each token is independently masked with probability $1 - \alpha_t$:
\begin{equation}
\begin{aligned}
q(\mathbf{x}_t | \mathbf{x}_0)
 &= \prod_{i=1}^L \Bigl[ \alpha_t \mathbb{I}(x_t^i = x_0^i) \\
 &\qquad + (1 - \alpha_t) \mathbb{I}(x_t^i = \texttt{[M]}) \Bigr],
\end{aligned}
\end{equation}
where $\texttt{[M]}$ denotes the mask token. As $t$ increases, the token-retention probability $\alpha_t$ decreases monotonically.
The reverse process learns to reconstruct the original data from the corrupted state. A neural network $p_\theta$ is trained to predict the original tokens for all masked positions simultaneously. The learning objective minimizes the negative log-likelihood over the masked indices $M_t$:
\begin{equation}
\mathcal{L}(\theta) = \mathbb{E}_{t, \mathbf{x}_0, \mathbf{x}_t} \left[ - \sum_{i \in M_t} \log p_\theta(x_0^i \mid \mathbf{x}_t) \right].
\end{equation}
\subsubsection{Confidence-Based Parallel Decoding}
The index $t$ in Eqs.~(1)--(2) denotes diffusion (noise) time. From here on we instead index the decoding trajectory: step $t=1,2,\dots$ takes the state $\mathbf{x}^{t-1}$ as input and produces $\mathbf{x}^{t}$, where $\mathbf{x}^{0}$ is the initial fully masked sequence and $M_t$ collects the positions still masked at step $t$.
To accelerate inference, Fast-dLLM  \citep{wu2025fast} employs a Confidence-Based Parallel Decoding (CBPD) strategy that iteratively fixes tokens based on predictive certainty. At each step $t$, given the current state $\mathbf{x}^{t-1}$, the model predicts the probability distribution over the vocabulary for all masked positions. CBPD identifies the set of high-confidence positions, denoted as $\mathcal{H}_t$, where the model's top prediction probability is at least $\phi$:
\begin{equation}
\mathcal{H}_t = \left\{ i \in M_t \;\middle|\; \max_{v} p_\theta(x_i=v \mid \mathbf{x}^{t-1}) \geq \phi \right\}.
\end{equation}
The state is then updated by unmasking these high-confidence tokens with their greedy predictions, while keeping uncertain positions masked for future refinement:
\begin{equation}
x^t_i = \begin{cases}
\operatorname*{arg\,max}_{v} p_\theta(x_i=v \mid \mathbf{x}^{t-1}) & \text{if } i \in \mathcal{H}_t. \\
x^{t-1}_i & \text{otherwise.}
\end{cases}
\end{equation}
If $\mathcal{H}_t$ is empty, CBPD  enforces an update on the single position with the highest global confidence to guarantee convergence.

\Needspace{5\baselineskip}
\subsection{Barriers to Parallel Decoding}
\label{section:Barriers to Parallel Decoding}

CBPD commits tokens whose marginal confidence exceeds a threshold. Fast-dLLM provides a sufficient high-confidence condition for greedy marginal modes to agree with the joint mode \citep{wu2025fast}; practical confidence thresholding is a conservative selection heuristic.

\begin{figure*}[!t]
  \centering
  \includegraphics[width=0.43\textwidth]{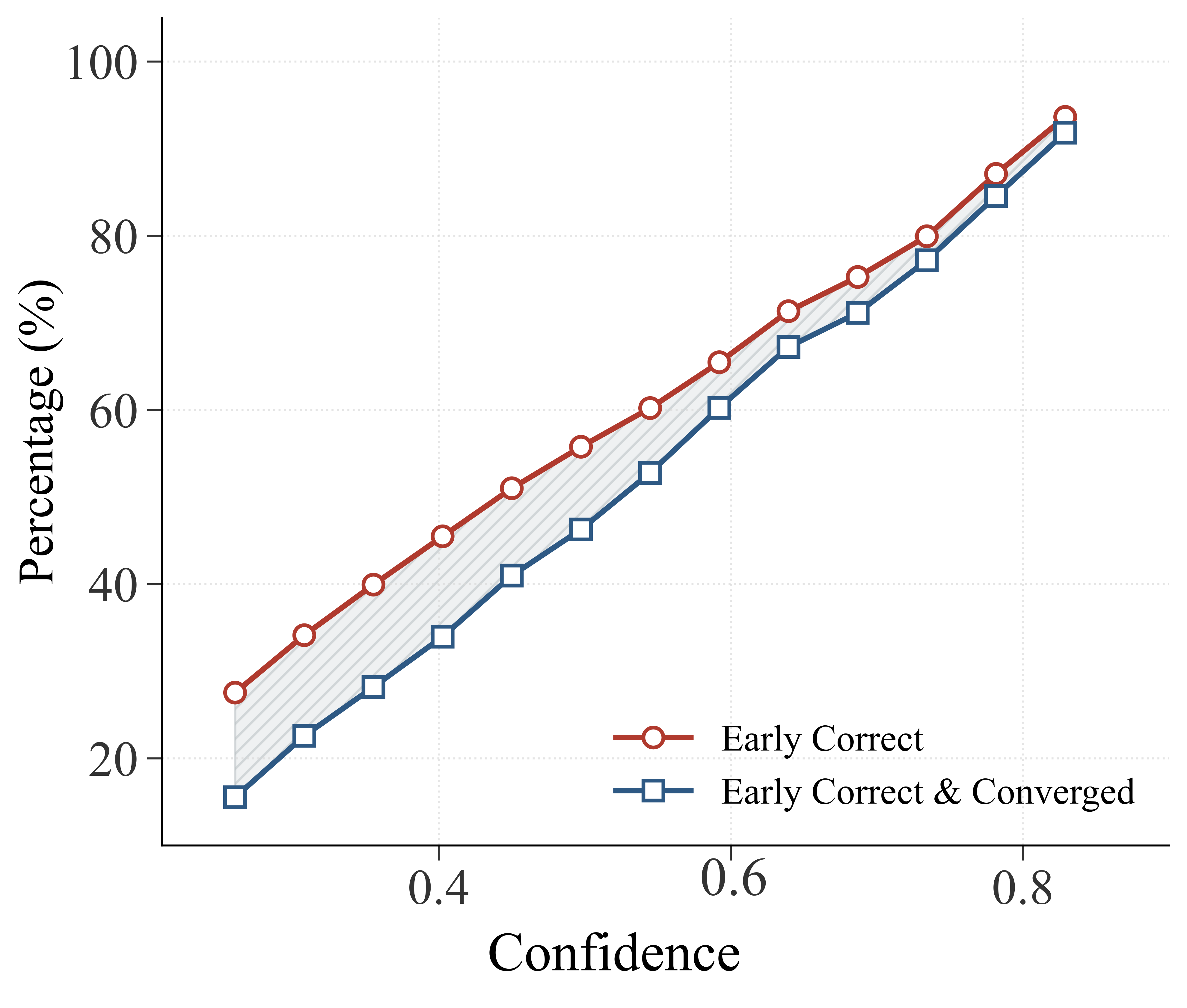}
  \hfill
  \includegraphics[width=0.53\textwidth]{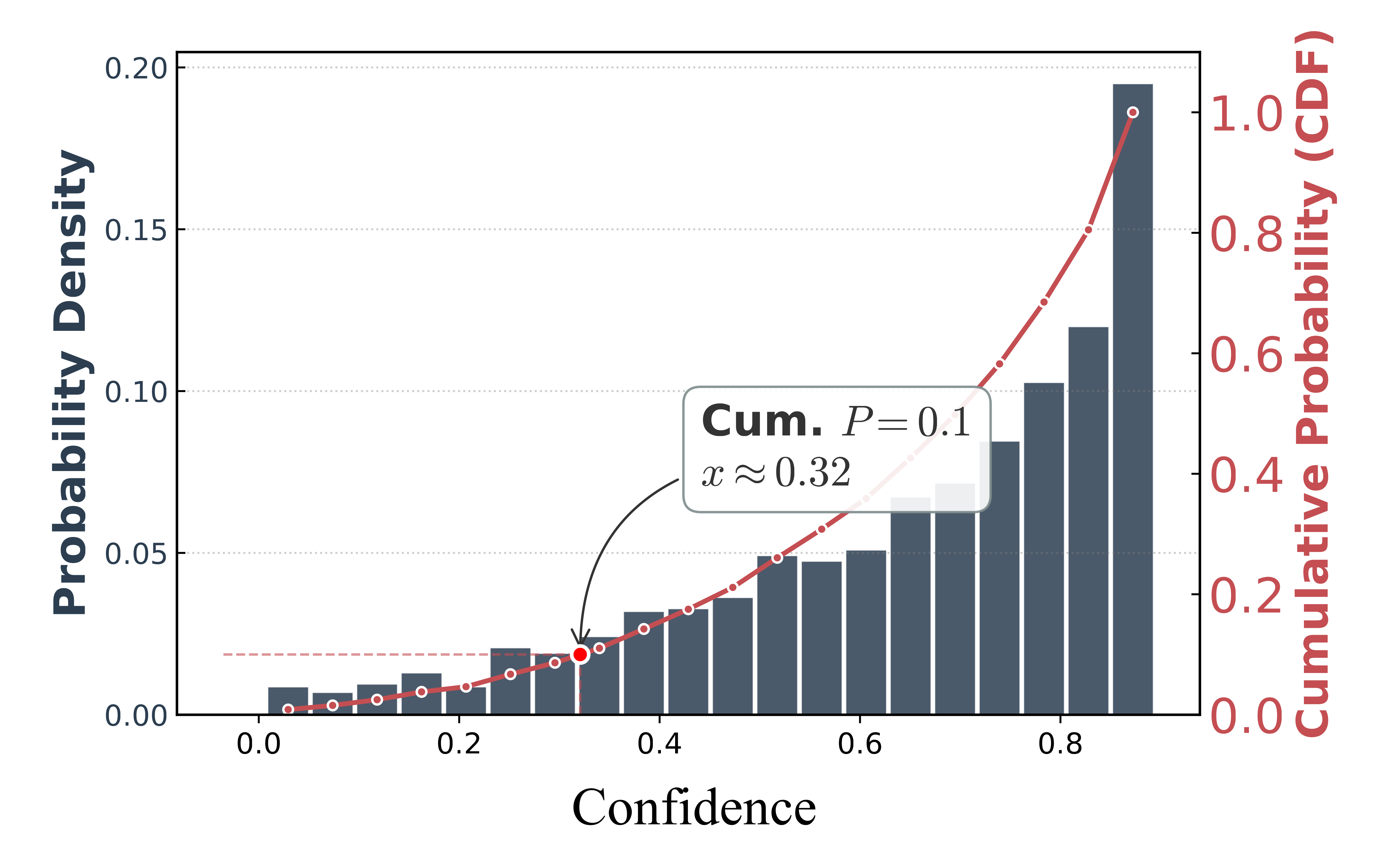}
  \caption{Left: fraction of tokens that are early-correct (red) and early-correct-and-converged (blue), as a function of confidence, on GSM8K with LLaDA-8B-Instruct. Right: preceding-step confidence of CBPD-reference tokens, shown as a histogram and CDF; only 10\% fall below 0.32.}
  \label{fig:convergence_confidence}
\end{figure*}

However, the strict confidence threshold in CBPD creates a bottleneck for the parallel decoding candidate set. In practice, we identify a large volume of `early correct' tokens—predictions that match the final high-confidence output before the threshold is met. As shown by the red line in Fig.~\ref{fig:convergence_confidence}~(left), more than half of the tokens in the medium confidence range ($[0.5, 0.9)$) are early correct. Furthermore, we observe `early correct and converged' tokens, which maintain stable and correct predictions throughout the pre-threshold phase. Fig.~\ref{fig:convergence_confidence}~(left) also illustrates the gap between these two categories across different confidence levels. At higher confidence levels($\ge0.6$), the curves exhibit a minimal gap, suggesting a high degree of overlap between these token subsets. This indicates that CBPD unnecessarily discards valid candidates. Naively reducing the threshold can commit predictions that change when dependencies are resolved, reducing accuracy (Table \ref{motivation_table}).

\begin{center}
\begin{minipage}{\textwidth}
\captionsetup{hypcap=false}
  \centering
  \begin{tabular}{cc}
    \toprule
    Threshold & HumanEval Acc \\
    \midrule
    0.9 & 42.1\% \\
    0.8 & 40.2\% \\
    0.7 & 36.0\% \\
    0.6 & 32.9\% \\
    \bottomrule
  \end{tabular}
  \captionof{table}{Performances of LLaDA-8B-Instruct with various confidence decoding thresholds.}
  \label{motivation_table}
\end{minipage}
\end{center}

\Needspace{5\baselineskip}
\subsection{Problem Formulation}
Formally, let $k_i$ denote the specific decoding step where the $i$-th token is decoded in CBPD. The prediction for this token is obtained by selecting the candidate with the highest probability, conditioned on the context $\mathbf{x}^{k_i-1}$ available at that step: 
\begin{equation}
\begin{aligned}
\pi(i\mid\mathbf{x}^{k_i-1}) &:= x_i^* \\
 &= \operatorname*{argmax}_{v \in \mathcal{V}} \,
 p_\theta\left(x_i = v \mid \mathbf{x}^{k_i-1}\right),
\end{aligned}
\end{equation}
where function $\pi$ represents greedy decoding. We take \( x_i^* \) obtained from the $k_i$-step prediction as the final convergence target. To maximize parallelism, iteration $t$ should decode all tokens that align with the final converged outcome. Formally, we define the maximal decodable set at step $t$ as:
\begin{equation}
    \mathcal{D}_t^* = \left\{ i \in M_t \;\middle|\; \pi(i \mid \mathbf{x}^{t-1}) = x_i^* \right\},
\end{equation}
where \( M_t \) is the set of candidate mask tokens in step $t$. This oracle set quantifies the parallel commitment opportunities that agree with the CBPD reference at each step.

However, the direct computation of $x_i^*$ at step $t$ is precluded by the unavailability of the future context $\mathbf{x}^{k_i-1}$. Thus, the challenge lies in maximizing the decodable set $\mathcal{D}_t$ at step $t$ while introducing minimal deviation from the original converged output.

\begin{figure*}[!t]
  \begin{centering}
        \centerline{\includegraphics[width=\textwidth]{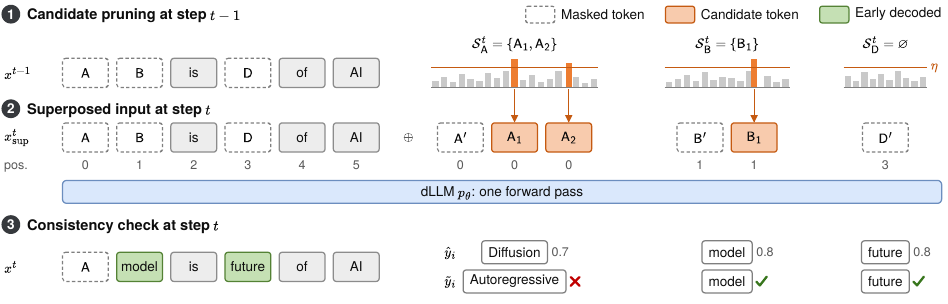}}
        \caption{Overview of LEAP. Candidates $\mathcal{S}_i^t$ are pruned from the step-$t-1$ predictions using $\eta$; $x^{t-1}$ is the updated state carried into step $t$. Candidates and copied masks reuse their original position IDs. One forward pass yields original predictions $\hat y_i$ and perturbed predictions $\widetilde{\hat y}_i$. With $\tau=0.7$ and $\phi=0.9$, B and D pass consistency and confidence checks, while A remains masked. D can pass even when its own candidate set is empty, because other positions supply the perturbation.}
    \label{overview}
  \end{centering}
\end{figure*}

\Needspace{5\baselineskip}
\subsection{Lookahead Early-Convergence Token Detection}
We propose LEAP, a parallel accelerated decoding strategy that identifies early converged tokens to unlock higher decoding parallelism. Fig. \ref{overview} shows an overview of the method.

Early convergence signifies stability under future context updates. As an ideal criterion, let $\Omega_i(\mathbf{x}^{t-1})$ contain the valid future contexts reachable by updating other masked positions while keeping position $i$ masked. An ideally converged token should satisfy:
\begin{equation} 
\forall \mathbf{c} \in \Omega_i(\mathbf{x}^{t-1}), \quad \pi(i \mid \mathbf{c}) = \pi(i \mid \mathbf{x}^{t-1}).
\end{equation}
LEAP detects stable predictions through a single aggregate perturbation constructed from pruned future-token candidates. Two design choices make this lookahead efficient: (1) candidate pruning uses the preceding step's predictions to retain plausible future context, and (2) multi-sequence superposition evaluates original and perturbed predictions within one forward pass. Algorithm~\ref{alg:leap} combines this consistency test with conventional high-confidence decoding to expand the committed set.

\textbf{Future Context Candidate Pruning.} Motivated by the `early correct' observation, we examine whether final CBPD-reference tokens already have non-negligible probability in the preceding step. Fig.~\ref{fig:convergence_confidence}~(right) shows that about 91\% have probability at least 0.3. This concentration of probability motivates a lower pruning threshold that retains plausible reference-token candidates and supplies informative context for the next decoding step. For $t\geq2$, step $t-1$ takes $\mathbf{x}^{t-2}$ as input, so its predictions define:
\begin{equation}
    \mathcal{S}_{i}^{t} = \left\{ v \in \mathcal{V} \mid p_{\theta}(x_i = v \mid \mathbf{x}^{t-2}) \geq \eta \right\}.
\end{equation}
We initialize $\mathcal{S}_i^1=\varnothing$; no preceding prediction exists at the first step.
\Needspace{19\baselineskip}
\begin{wrapfigure}{r}{0.38\textwidth}
\vspace{-8pt}
\captionsetup{font=small}
\centering
\includegraphics[width=\linewidth]{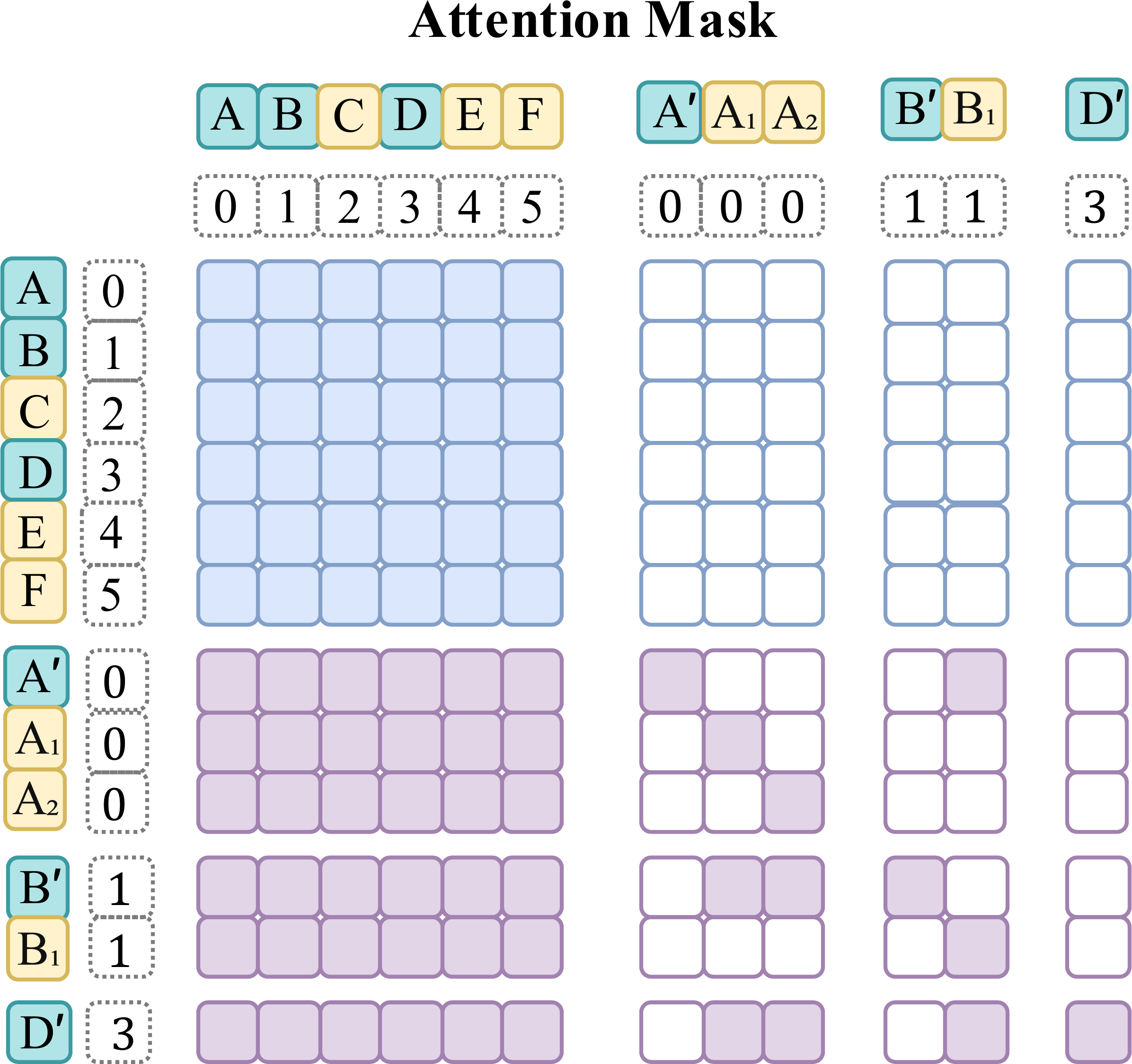}
\caption{Attention mask for separating the original and perturbed inference paths.}
\label{fig:attn_mask}
\end{wrapfigure}
\textbf{Multi-Sequence Superimposed Consistency Detection.} Early convergence indicates that a token remains stable under new context. Treating emerging context as a perturbation, we seek tokens that are robust to potential future variations. Evaluating every possible future context separately is intractable because the candidate combinations grow rapidly. We therefore approximate future perturbations by appending the pruned candidate tokens and copies of all masked tokens to the current sequence, while retaining their original position IDs to preserve contextual order.

The attention mask in Fig. \ref{fig:attn_mask} separates the original and perturbed inference paths. The original sequence cannot attend to appended tokens, preserving predictions under the original context. Each copied mask attends to the augmented context except for candidate tokens at its own position, which prevents direct answer leakage. A single forward pass thus yields predictions under both contexts; tokens that remain consistent across them are identified as early-converged. Formally, the superimposed sequence at step $t$ is:

\begin{equation}
    x^t_{\text{sup}} = x^{t-1} \oplus \bigoplus_{i \in M_t} \left( [x_i^{t-1}] \oplus \mathcal{S}^t_i \right),
\end{equation}
where $\oplus$ denotes concatenation. Let $\widetilde{\hat y}_i$ denote the greedy prediction read at the copied-mask row for position $i$ in this pass. The practical consistency set is:
\begin{equation}
    \mathcal{D}_t=\{ i\in M_t \mid \hat{y}_i = \widetilde{\hat y}_i \},
\end{equation}
where $\hat{y}_i = \pi(i \mid x^{t-1})$ is the original-context prediction. An unchanged top-1 token can still have low probability, so we also require its original-context confidence to reach $\tau$. When $\bigcup_{j\in M_t}\mathcal{S}_j^t$ is empty, we disable early decoding. Otherwise, the early-decoding set is:
\begin{equation}
    \hat{\mathcal{D}}_t = \left\{ i \in \mathcal{D}_t \mid p_\theta(x_i=\hat{y}_i \mid x^{t-1})\geq \tau\right\}.
\end{equation}
At step $t$, LEAP unmasks $\mathcal{U}_t=\mathcal{H}_t\cup\hat{\mathcal{D}}_t\cup\{g_t\}$, where $g_t=\arg\max_{i\in M_t}p_\theta(x_i=\hat y_i\mid x^{t-1})$. The high-confidence set uses the same $\phi$ as CBPD; the consistency test adds positions in $[\tau,\phi)$. The most confident position is always retained as the CBPD progress rule, even if it fails consistency.

\FloatBarrier
\Needspace{6\baselineskip}
\section{Experiments}
\Needspace{5\baselineskip}
\subsection{Experimental Setup}

\textbf{Benchmarks.} We evaluate LEAP on representative tasks across diverse domains: GSM8K \citep{cobbe2021training}, MATH \citep{hendrycks2021measuring}, and GPQA \citep{rein2024gpqa} for science and math reasoning; HumanEval \citep{chen2021evaluating} and MBPP \citep{austin2021program} for code generation. Furthermore, to demonstrate that our method is orthogonal to model-centric acceleration techniques, we integrate it with dParallel, a distillation-based method designed to enhance dLLM parallelism.

\textbf{Evaluation Metrics.} Accuracy (Acc) denotes benchmark-specific task accuracy and is reported as Pass@1 for HumanEval and MBPP. Tokens per second (TPS) measures end-to-end generation throughput, while tokens per forward pass (TPF) measures the average number of tokens committed per denoising forward pass. Steps denotes the total number of denoising iterations, and Spd(Lat.) is the full-step baseline latency divided by the method latency. TFOPs count token--forward-pass operations, whereas TFLOPS estimate total floating-point computation over the complete decoding trajectory.

\Needspace{29\baselineskip}
\begin{wrapfigure}{r}{0.55\textwidth}
\vspace{-8pt}
\small
\captionsetup{font=small,hypcap=false}
\hrule\vspace{4pt}
\refstepcounter{algorithm}\label{alg:leap}
  \captionof*{algorithm}{\textbf{Algorithm~\thealgorithm} LEAP parallel decoding.}
  \begin{algorithmic}[1]
    \REQUIRE Initial state $\mathbf{x}^{0}$, thresholds $\eta<\tau<\phi$
    \STATE $t \gets 1$;\ $\mathcal{S}_i^{1} \gets \varnothing$ for $i\in M_1$
    \WHILE{$M_t \neq \varnothing$}
      \STATE $\mathbf{x}_{\mathrm{sup}}^{t} \gets \operatorname{Superimpose}(\mathbf{x}^{t-1},\mathcal{S}^{t})$
      \STATE $(p_i,\widetilde p_i)_{i\in M_t} \gets p_\theta(\mathbf{x}_{\mathrm{sup}}^{t})$
      \STATE $\hat y_i \gets \arg\max_v p_i(v)$; $c_i \gets p_i(\hat y_i)$
      \STATE $\mathcal{H}_t \gets \{i\in M_t:c_i\geq\phi\}$ \hfill
      \STATE $\mathcal{E}_t \gets \{i\in M_t\setminus\mathcal{H}_t:\bigcup_j\mathcal{S}_j^{t}\neq\varnothing,\ \hat y_i=\arg\max_v\widetilde p_i(v),\ c_i\geq\tau\}$
      \STATE $g_t\gets\arg\max_{i\in M_t}c_i$
      \STATE $\mathcal{U}_t \gets \mathcal{H}_t\cup\mathcal{E}_t\cup\{g_t\}$
      \STATE $\mathbf{x}^{t} \gets \operatorname{Decode}(\mathbf{x}^{t-1},\mathcal{U}_t,\hat{\mathbf y})$
      \STATE $\mathcal{S}_i^{t+1} \gets \{v\in\mathcal{V}:p_i(v)\geq\eta\}$ for $i\in M_{t+1}$
      \STATE $t \gets t+1$
    \ENDWHILE
    \RETURN $\mathbf{x}^{t-1}$
  \end{algorithmic}
\vspace{4pt}\hrule

\end{wrapfigure}

\textbf{Baselines.} We compare LEAP against three decoding strategies: (1) \textit{Conf-Based}, confidence threshold-based parallel decoding \citep{wu2025fast} with the threshold fixed at $\phi=0.9$, (2) \textit{KLASS} \citep{kim2026klass} and (3) \textit{LoPA} \citep{xu2025lopa}.

\textbf{Implementation Details.} Our experiments are conducted on two open-source dLLMs: LLaDA-8B-Instruct and Dream-7B-Instruct. Both models utilize a semi-autoregressive generation strategy with a block size of 32. All methods use greedy sampling and a batch size of 1, with a generation length of 512 for the code benchmarks (HumanEval and MBPP) and 256 for the remaining ones. Unless otherwise specified, the hyperparameters are set to $\eta=0.2$ and $\tau=0.7$, and LEAP's safe-decoding threshold is $\phi=0.9$, matching the Conf-Based baseline. All experiments are performed on a single NVIDIA 5090 (32GB) GPU.

\begin{table*}[!t]
\centering
\small
\setlength{\tabcolsep}{5.2pt}

\begin{tabular}{llcccccccc}
\toprule
\multirow{2}{*}{Dataset}
& \multirow{2}{*}{Method}
& \multicolumn{4}{c}{\textbf{LLaDA-8B-Instruct}}
& \multicolumn{4}{c}{\textbf{Dream-7B-Instruct}} \\
\cmidrule(lr){3-6}
\cmidrule(lr){7-10}
&
& Acc$\uparrow$ & TPS$\uparrow$ & Steps$\downarrow$ & Spd(Lat.)$\uparrow$
& Acc$\uparrow$ & TPS$\uparrow$ & Steps$\downarrow$ & Spd(Lat.)$\uparrow$ \\
\midrule

\multirow{5}{*}{\shortstack[l]{GSM8K\\(4-shot)}}
& Baseline   & 76.3 & 8.4  & 256  & 1.0$\times$ & 78.2 & 5.7  & 256  & 1.0$\times$ \\
& LoPA       & --   & --   & --   & --           & 75.6 & 12.5 & 73   & 2.2$\times$ \\
& KLASS      & \textbf{78.8} & 15.2 & 115  & 1.8$\times$ & 76.1 & 12.3 & 113  & 2.2$\times$ \\
& Conf-Based & 76.5 & 27.2 & 79   & 3.2$\times$ & \textbf{77.6} & 24.7 & 60   & 4.2$\times$ \\
& \cellcolor{gray!15}LEAP & \cellcolor{gray!15}78.0 & \cellcolor{gray!15}\textbf{34.3} & \cellcolor{gray!15}\textbf{58} & \cellcolor{gray!15}\textbf{4.0}$\times$
               & \cellcolor{gray!15}76.9 & \cellcolor{gray!15}\textbf{31.3} & \cellcolor{gray!15}\textbf{44} & \cellcolor{gray!15}\textbf{5.5}$\times$ \\
\midrule

\multirow{5}{*}{\shortstack[l]{HumanEval\\(0-shot)}}
& Baseline   & 42.7 & 10.8 & 512  & 1.0$\times$ & 56.7 & 3.4  & 512  & 1.0$\times$ \\
& LoPA       & --   & --   & --   & --           & 52.4 & 12.4 & 76   & 4.0$\times$ \\
& KLASS      & \textbf{42.7} & 20.1 & 219  & 1.9$\times$ & \textbf{55.5} & 12.7 & 94   & 4.9$\times$ \\
& Conf-Based & 42.1 & 30.2 & 176  & 2.8$\times$ & 54.3 & 26.0 & 65   & 7.5$\times$ \\
& \cellcolor{gray!15}LEAP & \cellcolor{gray!15}\textbf{42.7} & \cellcolor{gray!15}\textbf{36.5} & \cellcolor{gray!15}\textbf{125} & \cellcolor{gray!15}\textbf{3.3}$\times$
               & \cellcolor{gray!15}54.3 & \cellcolor{gray!15}\textbf{33.9} & \cellcolor{gray!15}\textbf{50} & \cellcolor{gray!15}\textbf{8.8}$\times$ \\
\midrule

\multirow{5}{*}{\shortstack[l]{MBPP\\(3-shot)}}
& Baseline   & 37.0 & 1.0  & 512  & 1.0$\times$ & 55.0 & 1.0  & 512  & 1.0$\times$ \\
& LoPA       & --   & --   & --   & --           & 54.6 & 10.0 & 34   & 10.1$\times$ \\
& KLASS      & \textbf{40.2} & 15.2 & 67   & 7.6$\times$ & \textbf{59.6} & 9.2  & 54   & 9.1$\times$ \\
& Conf-Based & 36.2 & 11.9 & 45   & 11.3$\times$ & 54.8 & 17.8 & 27   & 18.1$\times$ \\
& \cellcolor{gray!15}LEAP & \cellcolor{gray!15}37.0 & \cellcolor{gray!15}\textbf{21.6} & \cellcolor{gray!15}\textbf{22} & \cellcolor{gray!15}\textbf{20.3}$\times$
               & \cellcolor{gray!15}53.0 & \cellcolor{gray!15}\textbf{25.2} & \cellcolor{gray!15}\textbf{18} & \cellcolor{gray!15}\textbf{25.0}$\times$ \\
\midrule

\multirow{5}{*}{\shortstack[l]{MATH\\(4-shot)}}
& Baseline   & 33.4 & 8.9  & 256  & 1.0$\times$ & 42.6 & 8.4  & 256  & 1.0$\times$ \\
& LoPA       & --   & --   & --   & --           & 41.7 & 11.2 & 107  & 1.3$\times$ \\
& KLASS      & 32.3 & 14.7 & 136  & 1.5$\times$ & 37.7 & 13.6 & 132  & 1.9$\times$ \\
& Conf-Based & \textbf{33.1} & 22.3 & 101  & 2.5$\times$ & \textbf{42.1} & 24.1 & 88   & 2.9$\times$ \\
& \cellcolor{gray!15}LEAP & \cellcolor{gray!15}32.4 & \cellcolor{gray!15}\textbf{27.6} & \cellcolor{gray!15}\textbf{73} & \cellcolor{gray!15}\textbf{3.1}$\times$
               & \cellcolor{gray!15}40.0 & \cellcolor{gray!15}\textbf{31.5} & \cellcolor{gray!15}\textbf{63} & \cellcolor{gray!15}\textbf{3.7}$\times$ \\
\midrule

\multirow{5}{*}{\shortstack[l]{GPQA\\(5-shot)}}
& Baseline   & 30.6 & 4.0  & 256  & 1.0$\times$ & 33.3 & 0.5  & 256  & 1.0$\times$ \\
& LoPA       & --   & --   & --   & --           & 30.6 & 7.6  & 6    & 20.7$\times$ \\
& KLASS      & 29.0 & 10.1 & 124  & 2.1$\times$ & \textbf{32.6} & 11.6 & 8    & 24.8$\times$ \\
& Conf-Based & 31.9 & 8.3  & 28   & 9.1$\times$ & 32.1 & \textbf{22.3} & 5    & 49.6$\times$ \\
& \cellcolor{gray!15}LEAP & \cellcolor{gray!15}\textbf{32.1} & \cellcolor{gray!15}\textbf{12.1} & \cellcolor{gray!15}\textbf{17} & \cellcolor{gray!15}\textbf{13.3}$\times$
               & \cellcolor{gray!15}32.4 & \cellcolor{gray!15}22.0 & \cellcolor{gray!15}\textbf{4} & \cellcolor{gray!15}\textbf{62.0}$\times$ \\

\bottomrule
\end{tabular}
\caption{
Main results on five benchmarks under two base models. ``--'' indicates no available implementation for that model. Bold marks the best value among the accelerated methods; the full-step Baseline is excluded.
}
\label{tab:main_results_two_models}
\end{table*}

\Needspace{5\baselineskip}
\subsection{Main Results}

\textbf{Results on LLaDA and Dream.} Table \ref{tab:main_results_two_models} shows that LEAP achieves the highest decoding speed on most benchmarks. On LLaDA-8B-Instruct, its average speedup (arithmetic mean over the five benchmarks) is $8.8\times$, compared with $5.8\times$ for Conf-Based and $3.0\times$ for KLASS; average accuracy is 44.4\%, versus 44.0\% for the baseline and 44.0\% for Conf-Based. On Dream-7B-Instruct, LEAP reaches $21.0\times$ average speedup, compared with $16.5\times$ for Conf-Based, $8.6\times$ for KLASS, and $7.7\times$ for LoPA, while retaining comparable accuracy. Across both models, lower step counts show that early-convergence detection expands the decodable set per step. The consistency test preserves task accuracy while expanding the decodable set at a looser threshold: on HumanEval, plain confidence decoding at 0.7 reaches 36.0\% (Table~\ref{motivation_table}), while LEAP with $\tau=0.7$ reaches 42.7\% in fewer steps than the $\phi=0.9$ baseline.

\begin{figure*}[!t]
  \centering
  \includegraphics[width=0.485\textwidth]{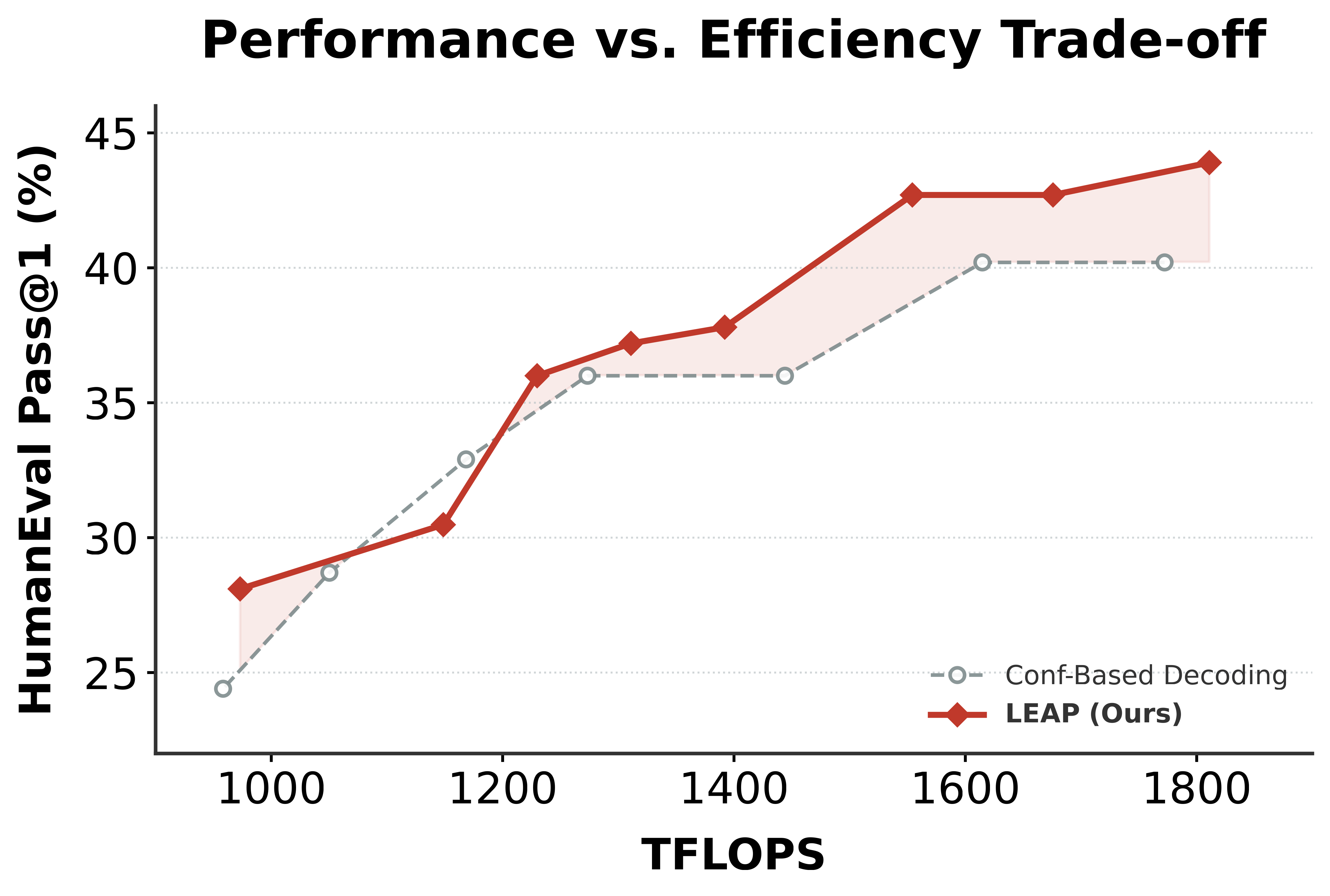}
  \hfill
  \includegraphics[width=0.485\textwidth]{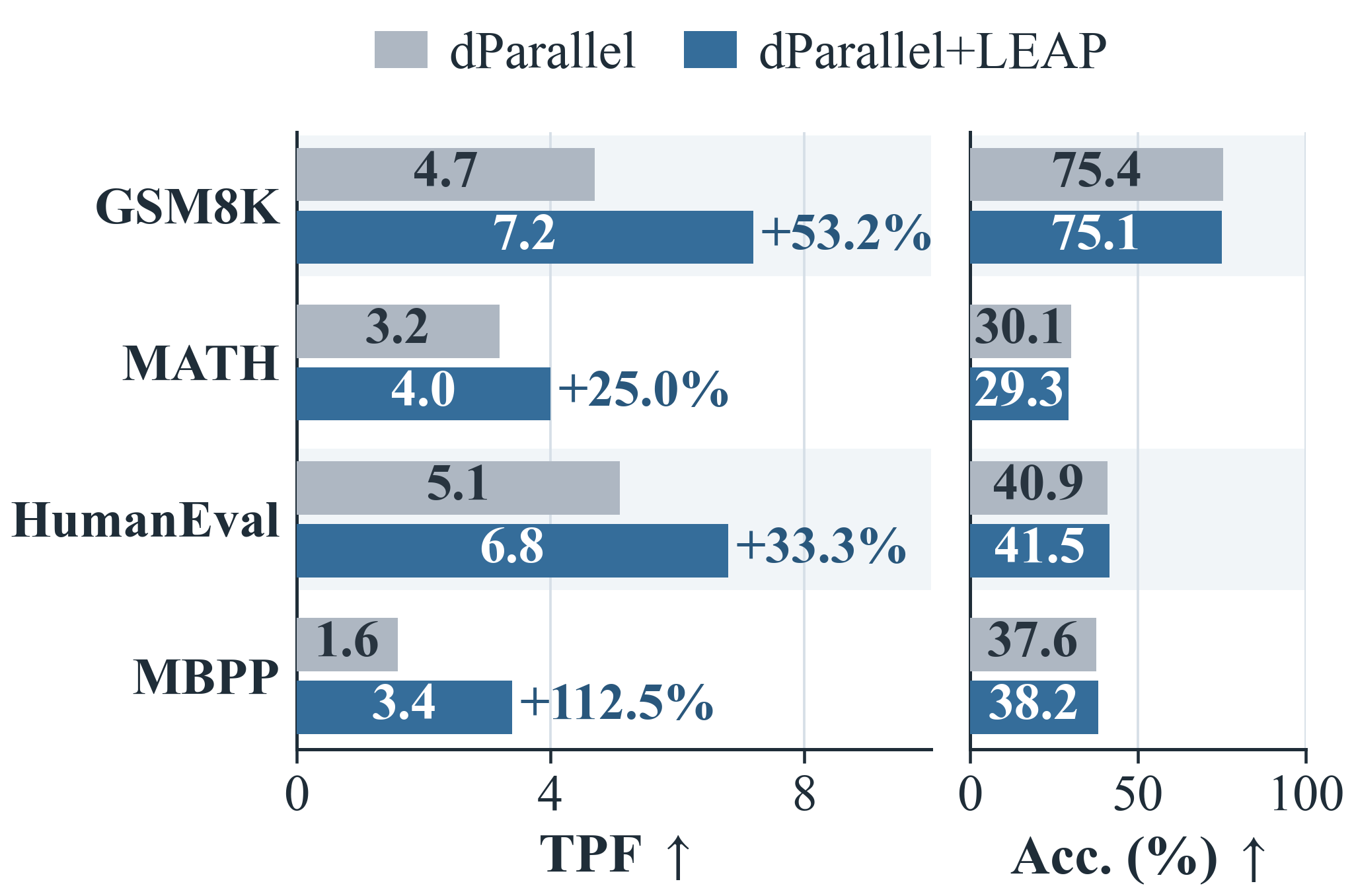}
  \caption{HumanEval efficiency--accuracy trade-off (left) and dParallel results on LLaDA-8B-Instruct (right).}
  \label{fig:efficiency_and_dparallel}
\end{figure*}

\textbf{Performance vs. Efficiency Analysis.} Figure~\ref{fig:efficiency_and_dparallel} (left) compares HumanEval Pass@1 with cumulative computation in TFLOPS. LEAP establishes a stronger accuracy--computation frontier in the medium-to-high Pass@1 range: it reaches comparable accuracy with less computation and delivers higher accuracy at comparable budgets. This improvement extends the benefits of increased decoding parallelism to the complete generation trajectory.

Lookahead consistency identifies stable predictions before they reach a conservative confidence threshold, reducing iterations spent repeatedly predicting these positions. The TFLOPS comparison includes the computation of superimposed inputs and shows that fewer denoising passes outweigh their additional per-pass work. Together with the higher TPS and latency speedups in Table~\ref{tab:main_results_two_models}, the frontier demonstrates gains in both computational efficiency and practical generation speed.

\textbf{Results on dParallel.} We integrate LEAP with dParallel-distilled LLaDA-8B-Instruct on mathematical reasoning and code generation tasks. dParallel's certainty-forcing distillation makes masked-token predictions reach high confidence earlier and in parallel \citep{chen2025dparallel}. LEAP complements this model-level acceleration by committing medium-confidence predictions that remain consistent under future-context perturbation, expanding the decodable set while using the distilled checkpoint directly.

Figure~\ref{fig:efficiency_and_dparallel} (right) shows that average TPF increases from 3.7 to 5.4 (46\%), with accuracy changes of at most 0.8 percentage points. On GSM8K, TPF rises from 4.7 to 7.2, with accuracy at 75.4\% and 75.1\%. These gains show that early-commitment opportunities persist after distillation improves confidence formation. LEAP converts them into additional parallelism through training-free, inference-time consistency selection, demonstrating the complementarity of model optimization and decoding optimization.

\begin{center}
\begin{minipage}{\textwidth}
\captionsetup{hypcap=false}
\centering
\small
\setlength{\tabcolsep}{3.2pt}
\renewcommand{\arraystretch}{1.06}
\begin{tabular}{lccccc}
\toprule
& \multicolumn{2}{c}{Conf-Based} & \multicolumn{2}{c}{LEAP} & \\
\cmidrule(lr){2-3}\cmidrule(lr){4-5}
Dataset & Acc. (\%) & TPF & Acc. (\%) & TPF & $\Delta$TPF \\
\midrule
GSM8K     & 91.0 & 1.8 & 90.4 & 2.4 & $+33.3\%$ \\
MATH      & 66.3 & 2.5 & 65.7 & 3.1 & $+24.0\%$ \\
MBPP      & 69.2 & 2.4 & 67.0 & 3.1 & $+29.2\%$ \\
HumanEval & 86.6 & 2.3 & 86.6 & 3.0 & $+30.4\%$ \\
\bottomrule
\end{tabular}
\captionof{table}{Results on LLaDA2.0-mini. $\Delta$TPF is relative to Conf-Based.}
\label{tab:llada2_results}
\end{minipage}
\end{center}

\textbf{Results on LLaDA2.0-mini.} We evaluate LEAP on the MoE-based LLaDA2.0-mini \citep{bie2025llada2} with KV-cache reuse enabled. With a generation length of 2,048 and a block size of 32, Conf-Based uses $\phi=0.95$, while LEAP uses $\tau=0.8$ and $\eta=0.2$. Table \ref{tab:llada2_results} shows consistent TPF gains across all four benchmarks (29.2\% on average) with modest accuracy changes, indicating that LEAP remains effective alongside MoE routing and KV-cache reuse.

\begin{figure}[!htbp]
  \centering
  \includegraphics[width=0.78\textwidth]{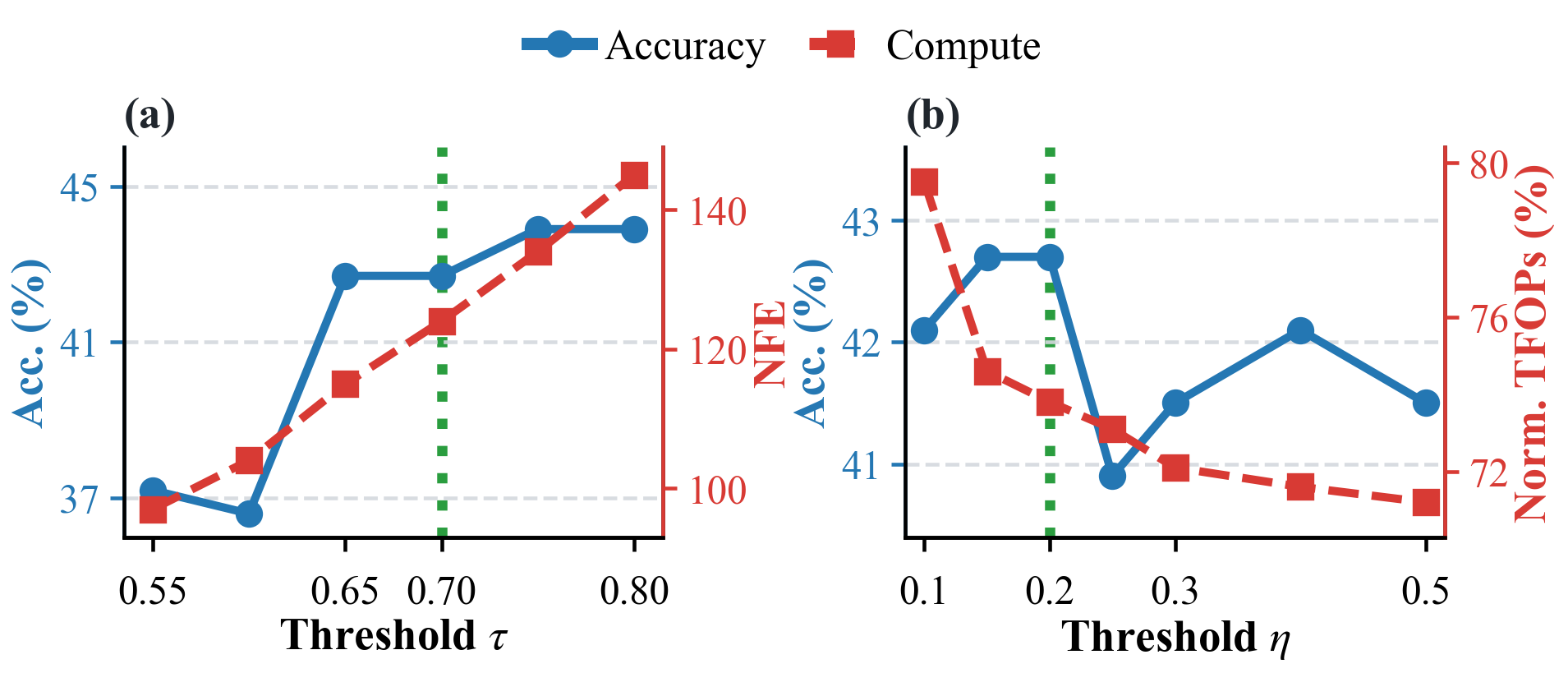}
  \caption{Hyperparameter sensitivity on HumanEval. Accuracy and NFE versus $\tau$ (a); accuracy and normalized TFOPs versus $\eta$ (b). Green lines mark $\tau=0.7$ and $\eta=0.2$.}
  \label{fig:four_images}
\end{figure}

\Needspace{5\baselineskip}
\subsection{Hyperparameter Analysis}
\textbf{Impact of threshold $\tau$.} We vary $\tau$ from 0.55 to 0.8 on HumanEval with LLaDA-8B-Instruct. Figure~\ref{fig:four_images}(a) shows that accuracy rises sharply at $\tau=0.65$ and then reaches a plateau; raising $\tau$ from 0.7 to 0.8 adds 1.2 accuracy points with 17\% more steps. GSM8K peaks at $\tau=0.7$, reaching 78.0\% in 58 steps, compared with 76.4\% in 66 steps at 0.8 (Appendix~\ref{app:sweeps}). These trends motivate $\tau=0.7$ as a balanced setting for accuracy and parallelism.

At the same threshold 0.7, LEAP reaches 42.7\% on HumanEval versus 36.0\% for confidence-based decoding, and 78.0\% on GSM8K versus 76.0\% (Table~\ref{tab:tau}). The gains of 6.7 and 2.0 points demonstrate effective consistency selection from the medium-confidence pool. Relative to CBPD at $\phi=0.9$, LEAP also reduces steps from 176 to 125 on HumanEval and from 79 to 58 on GSM8K, combining higher parallelism with preserved task performance.

\textbf{Impact of threshold $\eta$.} Token Forward Operations (TFOPs) count token--forward-pass operations summed across timesteps. We vary $\eta$ from 0.1 to 0.5 on HumanEval, normalizing TFOPs to confidence-based decoding. Figure~\ref{fig:four_images}(b) shows that lowering $\eta$ increases computation. LEAP still uses 20\%--30\% fewer TFOPs through fewer timesteps. We use $\eta=0.2$ to retain a compact pool of future-context candidates; Appendix~\ref{app:sweeps} reports the full sweep.

\begin{figure}[!htbp]
  \centering
  \includegraphics[width=0.96\textwidth]{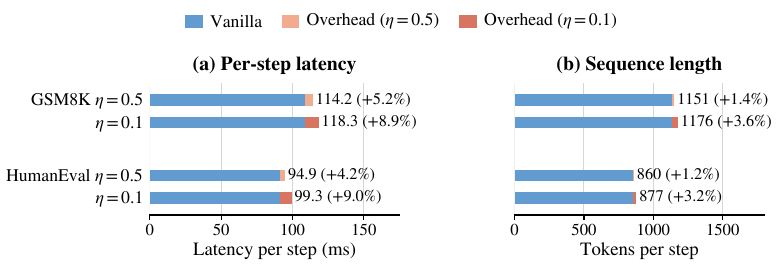}
  \caption{Per-step latency and sequence-length overhead on LLaDA-8B-Instruct. Blue: Vanilla; red: LEAP overhead for $\eta\in\{0.5,0.1\}$.}
  \label{fig:overhead_analysis}
\end{figure}

\Needspace{5\baselineskip}
\subsection{Per-Step Overhead Analysis}
\label{sec:overhead}
Figure~\ref{fig:overhead_analysis} shows how pruning keeps lookahead computation compact. Each retained candidate has probability at least $\eta$, so normalization gives $|\mathcal{S}_i^t|\leq\lfloor1/\eta\rfloor$. With $\eta=0.2$, each masked position contributes at most five candidates and one copied mask; a block of size 32 therefore appends at most 192 tokens. This bound is independent of vocabulary size and original context length. The shrinking masked set further reduces the number of probes and candidate tokens during decoding.

On HumanEval and GSM8K, average sequence length increases by only 1.2\%--3.6\%, and latency per step by 4.2\%--9.0\% (Table~\ref{tab:overhead}). Candidate pruning concentrates lookahead work on this small input extension. By decoding more tokens together, LEAP reduces denoising iterations and offsets the modest per-step cost. The resulting 20\%--30\% reduction in total TFOPs captures savings across the full trajectory; the TFLOPS frontier and measured latency demonstrate lower overall computation and faster generation. Compact perturbations and earlier token commitment together make LEAP's lookahead efficient.

\section{Conclusion}
In this paper, we introduce Lookahead Early-Convergence Token Detection for Accelerated Parallel Decoding (LEAP). By integrating future context candidate pruning with multi-sequence superimposed consistency detection, LEAP detects and decodes early converged tokens at a low computational cost. Empirically, LEAP improves model parallelism and decreases inference latency while largely preserving accuracy, and it alleviates the strict reliance on high-confidence decoding of dLLM parallel decoding, opening new avenues for parallel inference research.
\label{sec:content-end}

\begingroup
\small
\setlength{\bibsep}{2pt}
\bibliographystyle{plainnat}
\bibliography{main}

\begin{thebibliography}{27}
\providecommand{\natexlab}[1]{#1}
\providecommand{\url}[1]{\texttt{#1}}
\expandafter\ifx\csname urlstyle\endcsname\relax
  \providecommand{\doi}[1]{doi: #1}\else
  \providecommand{\doi}{doi: \begingroup \urlstyle{rm}\Url}\fi

\bibitem[Austin et~al.(2021)Austin, Odena, Nye, Bosma, Michalewski, Dohan,
  Jiang, Cai, Terry, Le, et~al.]{austin2021program}
Jacob Austin, Augustus Odena, Maxwell Nye, Maarten Bosma, Henryk Michalewski,
  David Dohan, Ellen Jiang, Carrie Cai, Michael Terry, Quoc Le, et~al.
\newblock {Program Synthesis with Large Language Models}.
\newblock \emph{arXiv preprint arXiv:2108.07732}, 2021.
\newblock \doi{10.48550/arXiv.2108.07732}.
\newblock URL \url{https://arxiv.org/abs/2108.07732}.

\bibitem[Ben-Hamu et~al.(2025)Ben-Hamu, Gat, Severo, Nolte, and
  Karrer]{ben2025accelerated}
Heli Ben-Hamu, Itai Gat, Daniel Severo, Niklas~S. Nolte, and Brian Karrer.
\newblock {Accelerated Sampling from Masked Diffusion Models via Entropy
  Bounded Unmasking}.
\newblock In \emph{Advances in Neural Information Processing Systems},
  volume~38, 2025.
\newblock \doi{10.52202/085713-1874}.
\newblock URL
  \url{https://proceedings.nips.cc/paper_files/paper/2025/hash/510e22f4a2da5212a64f1591736e2eaf-Abstract-Conference.html}.

\bibitem[Bie et~al.(2025)Bie, Cao, Chen, Du, Gong, Gong, Gu, Hu, Huang, Lan,
  et~al.]{bie2025llada2}
Tiwei Bie, Maosong Cao, Kun Chen, Lun Du, Mingliang Gong, Zhuochen Gong, Yanmei
  Gu, Jiaqi Hu, Zenan Huang, Zhenzhong Lan, et~al.
\newblock {LLaDA2.0: Scaling Up Diffusion Language Models to 100B}.
\newblock \emph{arXiv preprint arXiv:2512.15745}, 2025.
\newblock \doi{10.48550/arXiv.2512.15745}.
\newblock URL \url{https://arxiv.org/abs/2512.15745}.

\bibitem[Chen et~al.(2021)Chen, Tworek, Jun, Yuan, Pinto, Kaplan, Edwards,
  Burda, Joseph, Brockman, et~al.]{chen2021evaluating}
Mark Chen, Jerry Tworek, Heewoo Jun, Qiming Yuan, Henrique Ponde De~Oliveira
  Pinto, Jared Kaplan, Harri Edwards, Yuri Burda, Nicholas Joseph, Greg
  Brockman, et~al.
\newblock {Evaluating Large Language Models Trained on Code}.
\newblock \emph{arXiv preprint arXiv:2107.03374}, 2021.
\newblock \doi{10.48550/arXiv.2107.03374}.
\newblock URL \url{https://arxiv.org/abs/2107.03374}.

\bibitem[Chen et~al.(2026)Chen, Fang, Ma, Yu, and Wang]{chen2025dparallel}
Zigeng Chen, Gongfan Fang, Xinyin Ma, Ruonan Yu, and Xinchao Wang.
\newblock {dParallel: Learnable Parallel Decoding for dLLMs}.
\newblock In \emph{International Conference on Learning Representations}, 2026.
\newblock URL
  \url{https://proceedings.iclr.cc/paper_files/paper/2026/hash/57250222014c35949476f3f272c322d2-Abstract-Conference.html}.

\bibitem[Cobbe et~al.(2021)Cobbe, Kosaraju, Bavarian, Chen, Jun, Kaiser,
  Plappert, Tworek, Hilton, Nakano, et~al.]{cobbe2021training}
Karl Cobbe, Vineet Kosaraju, Mohammad Bavarian, Mark Chen, Heewoo Jun, Lukasz
  Kaiser, Matthias Plappert, Jerry Tworek, Jacob Hilton, Reiichiro Nakano,
  et~al.
\newblock {Training Verifiers to Solve Math Word Problems}.
\newblock \emph{arXiv preprint arXiv:2110.14168}, 2021.
\newblock \doi{10.48550/arXiv.2110.14168}.
\newblock URL \url{https://arxiv.org/abs/2110.14168}.

\bibitem[{DeepSeek-AI}(2024)]{liu2024deepseek}
{DeepSeek-AI}.
\newblock {DeepSeek-V3 Technical Report}.
\newblock \emph{arXiv preprint arXiv:2412.19437}, 2024.
\newblock \doi{10.48550/arXiv.2412.19437}.
\newblock URL \url{https://arxiv.org/abs/2412.19437}.

\bibitem[Fu et~al.(2026)Fu, Huang, Adams, Wang, Yi, Kamani, Srinivasan, and
  Jiao]{fu2025bits}
Hengyu Fu, Baihe Huang, Virginia Adams, Charles Wang, Junkeun Yi,
  Mohammad~Mahdi Kamani, Venkat~Krishna Srinivasan, and Jiantao Jiao.
\newblock {From Bits to Rounds: Parallel Decoding with Exploration for
  Diffusion Language Models}.
\newblock In \emph{Proceedings of the 43rd International Conference on Machine
  Learning}, volume 306 of \emph{Proceedings of Machine Learning Research},
  pages 31877--31896. PMLR, 2026.
\newblock URL \url{https://proceedings.mlr.press/v306/fu26h.html}.

\bibitem[{Google DeepMind}(2025)]{deepmind2025geminidiffusion}
{Google DeepMind}.
\newblock {Gemini Diffusion}.
\newblock \url{https://deepmind.google/models/gemini-diffusion/}, 2025.
\newblock Experimental research model.

\bibitem[Grattafiori et~al.(2024)Grattafiori, Dubey, Jauhri, Pandey, Kadian,
  Al-Dahle, Letman, Mathur, Schelten, Vaughan, et~al.]{dubey2024llama}
Aaron Grattafiori, Abhimanyu Dubey, Abhinav Jauhri, Abhinav Pandey, Abhishek
  Kadian, Ahmad Al-Dahle, Aiesha Letman, Akhil Mathur, Alan Schelten, Alex
  Vaughan, et~al.
\newblock {The Llama 3 Herd of Models}.
\newblock \emph{arXiv e-prints}, 2024.
\newblock \doi{10.48550/arXiv.2407.21783}.
\newblock URL \url{https://arxiv.org/abs/2407.21783v3}.
\newblock Version 3.

\bibitem[Hendrycks et~al.(2021)Hendrycks, Burns, Kadavath, Arora, Basart, Tang,
  Song, and Steinhardt]{hendrycks2021measuring}
Dan Hendrycks, Collin Burns, Saurav Kadavath, Akul Arora, Steven Basart, Eric
  Tang, Dawn Song, and Jacob Steinhardt.
\newblock {Measuring Mathematical Problem Solving With the MATH Dataset}.
\newblock In \emph{Proceedings of the Neural Information Processing Systems
  Track on Datasets and Benchmarks}, volume~1, 2021.
\newblock URL
  \url{https://datasets-benchmarks-proceedings.neurips.cc/paper/2021/hash/be83ab3ecd0db773eb2dc1b0a17836a1-Abstract-round2.html}.

\bibitem[{Inception Labs} et~al.(2025){Inception Labs}, Khanna, Kharbanda, Li,
  Varma, Wang, Birnbaum, Luo, Miraoui, Palrecha, Ermon, Grover, and
  Kuleshov]{inception2025mercury}
{Inception Labs}, Samar Khanna, Siddhant Kharbanda, Shufan Li, Harshit Varma,
  Eric Wang, Sawyer Birnbaum, Ziyang Luo, Yanis Miraoui, Akash Palrecha,
  Stefano Ermon, Aditya Grover, and Volodymyr Kuleshov.
\newblock {Mercury: Ultra-Fast Language Models Based on Diffusion}.
\newblock \emph{arXiv preprint arXiv:2506.17298}, 2025.
\newblock \doi{10.48550/arXiv.2506.17298}.
\newblock URL \url{https://arxiv.org/abs/2506.17298}.

\bibitem[Kim et~al.(2025)Kim, Hong, Jung, Park, and Yun]{kim2026klass}
Seo~Hyun Kim, Sunwoo Hong, Hojung Jung, Youngrok Park, and Se-Young Yun.
\newblock {KLASS: KL-Guided Fast Inference in Masked Diffusion Models}.
\newblock In \emph{Advances in Neural Information Processing Systems},
  volume~38, pages 102436--102470, 2025.
\newblock \doi{10.52202/085713-3087}.
\newblock URL
  \url{https://proceedings.neurips.cc/paper_files/paper/2025/hash/8514c1f12dcdd061ac8521feb78f28ca-Abstract-Conference.html}.

\bibitem[Li et~al.(2026)Li, Guan, Wu, and Li]{li2025refusion}
Jia-Nan Li, Jian Guan, Wei Wu, and Chongxuan Li.
\newblock {ReFusion: A Diffusion Large Language Model with Parallel
  Autoregressive Decoding}.
\newblock In \emph{International Conference on Learning Representations}, 2026.
\newblock URL
  \url{https://proceedings.iclr.cc/paper_files/paper/2026/hash/585979c057a1b30796cf317063559638-Abstract-Conference.html}.

\bibitem[Li et~al.(2025)Li, Zhou, Muhtar, Yin, Yan, Shen, Vosoughi, and
  Liu]{li2025diffusion}
Pengxiang Li, Yefan Zhou, Dilxat Muhtar, Lu~Yin, Shilin Yan, Li~Shen, Soroush
  Vosoughi, and Shiwei Liu.
\newblock {Diffusion Language Models Know the Answer Before Decoding}.
\newblock \emph{arXiv preprint arXiv:2508.19982}, 2025.
\newblock \doi{10.48550/arXiv.2508.19982}.
\newblock URL \url{https://arxiv.org/abs/2508.19982v5}.
\newblock Version 5.

\bibitem[Liu et~al.(2026{\natexlab{a}})Liu, He, Zeng, Zhang, Zhang, Wu, Jia,
  Liu, Zhou, and Zhou]{liu2025wedlm}
Aiwei Liu, Minghua He, Shaoxun Zeng, Sijun Zhang, Linhao Zhang, Chuhan Wu, Wei
  Jia, Yuan Liu, Xiao Zhou, and Jie Zhou.
\newblock {WeDLM: Reconciling Diffusion Language Models with Standard Causal
  Attention for Fast Inference}.
\newblock In \emph{Proceedings of the 43rd International Conference on Machine
  Learning}, volume 306 of \emph{Proceedings of Machine Learning Research},
  pages 76472--76489. PMLR, 2026{\natexlab{a}}.
\newblock URL \url{https://proceedings.mlr.press/v306/liu26av.html}.

\bibitem[Liu et~al.(2026{\natexlab{b}})Liu, Dong, Ye, Mehta, Fu, Singh, Kautz,
  Zhang, and Molchanov]{liu2025tidar}
Jingyu Liu, Xin Dong, Zhifan Ye, Rishabh Mehta, Yonggan Fu, Vartika Singh, Jan
  Kautz, Ce~Zhang, and Pavlo Molchanov.
\newblock {TiDAR: Think in Diffusion, Talk in Autoregression}.
\newblock In \emph{Proceedings of Machine Learning and Systems}, volume~8,
  2026{\natexlab{b}}.
\newblock URL
  \url{https://proceedings.mlsys.org/paper_files/paper/2026/hash/1367d856028f65a9555b0274db09e608-Abstract-Conference.html}.

\bibitem[Ma et~al.(2025)Ma, Yu, Fang, and Wang]{ma2025dkv}
Xinyin Ma, Runpeng Yu, Gongfan Fang, and Xinchao Wang.
\newblock {dKV-Cache: The Cache for Diffusion Language Models}.
\newblock In \emph{Advances in Neural Information Processing Systems},
  volume~38, pages 165470--165494, 2025.
\newblock \doi{10.52202/085713-4984}.
\newblock URL
  \url{https://proceedings.neurips.cc/paper_files/paper/2025/hash/db0ee27cb50dd9087b133f6e7d28a90e-Abstract-Conference.html}.

\bibitem[Nie et~al.(2025)Nie, Zhu, You, Zhang, Ou, Hu, Zhou, Lin, Wen, and
  Li]{nie2025large}
Shen Nie, Fengqi Zhu, Zebin You, Xiaolu Zhang, Jingyang Ou, Jun Hu, Jun Zhou,
  Yankai Lin, Ji-Rong Wen, and Chongxuan Li.
\newblock {Large Language Diffusion Models}.
\newblock In \emph{Advances in Neural Information Processing Systems},
  volume~38, 2025.
\newblock \doi{10.52202/085713-1689}.
\newblock URL
  \url{https://proceedings.neurips.cc/paper_files/paper/2025/hash/48b383b24230e0e6e649d9c98dae4d8c-Abstract-Conference.html}.

\bibitem[{OpenAI}(2023)]{achiam2023gpt}
{OpenAI}.
\newblock {GPT-4 Technical Report}.
\newblock \emph{arXiv preprint arXiv:2303.08774}, 2023.
\newblock \doi{10.48550/arXiv.2303.08774}.
\newblock URL \url{https://arxiv.org/abs/2303.08774}.

\bibitem[Rein et~al.(2024)Rein, Hou, Stickland, Petty, Pang, Dirani, Michael,
  and Bowman]{rein2024gpqa}
David Rein, Betty~Li Hou, Asa~Cooper Stickland, Jackson Petty, Richard~Yuanzhe
  Pang, Julien Dirani, Julian Michael, and Samuel~R Bowman.
\newblock {GPQA: A Graduate-Level Google-Proof Q\&A Benchmark}.
\newblock In \emph{First Conference on Language Modeling}, 2024.
\newblock URL \url{https://openreview.net/forum?id=Ti67584b98}.

\bibitem[Song et~al.(2025)Song, Zhang, Luo, Gao, Xia, Luo, Li, Yang, Yu, Qu,
  et~al.]{song2025seed}
Yuxuan Song, Zheng Zhang, Cheng Luo, Pengyang Gao, Fan Xia, Hao Luo, Zheng Li,
  Yuehang Yang, Hongli Yu, Xingwei Qu, et~al.
\newblock {Seed Diffusion: A Large-Scale Diffusion Language Model with
  High-Speed Inference}.
\newblock \emph{arXiv preprint arXiv:2508.02193}, 2025.
\newblock \doi{10.48550/arXiv.2508.02193}.
\newblock URL \url{https://arxiv.org/abs/2508.02193}.

\bibitem[Wang et~al.(2026)Wang, Xu, Jin, Jin, Zhang, Yu, and
  Deng]{wang2025diffusion}
Xu~Wang, Chenkai Xu, Yijie Jin, Jiachun Jin, Hao Zhang, Kai Yu, and Zhijie
  Deng.
\newblock {Diffusion LLMs Can Do Faster-Than-AR Inference via Discrete
  Diffusion Forcing}.
\newblock In \emph{International Conference on Learning Representations}, 2026.
\newblock URL
  \url{https://proceedings.iclr.cc/paper_files/paper/2026/hash/04bb76a98d9f32226b3beba7bd26a51f-Abstract-Conference.html}.

\bibitem[Wu et~al.(2026)Wu, Zhang, Xue, Liu, Diao, Zhu, Luo, Han, and
  Xie]{wu2025fast}
Chengyue Wu, Hao Zhang, Shuchen Xue, Zhijian Liu, Shizhe Diao, Ligeng Zhu, Ping
  Luo, Song Han, and Enze Xie.
\newblock {Fast-dLLM: Training-free Acceleration of Diffusion LLM by Enabling
  KV Cache and Parallel Decoding}.
\newblock In \emph{International Conference on Learning Representations}, 2026.
\newblock URL
  \url{https://proceedings.iclr.cc/paper_files/paper/2026/hash/5d8d4e6061c3ba96c240b7fa1ae3471d-Abstract-Conference.html}.

\bibitem[Xu et~al.(2025)Xu, Jin, Li, Tu, Long, Tu, Song, Si, Hou, Yan,
  et~al.]{xu2025lopa}
Chenkai Xu, Yijie Jin, Jiajun Li, Yi~Tu, Guoping Long, Dandan Tu, Mingcong
  Song, Hongjie Si, Tianqi Hou, Junchi Yan, et~al.
\newblock {LoPA: Scaling dLLM Inference via Lookahead Parallel Decoding}.
\newblock \emph{arXiv preprint arXiv:2512.16229}, 2025.
\newblock \doi{10.48550/arXiv.2512.16229}.
\newblock URL \url{https://arxiv.org/abs/2512.16229}.

\bibitem[Yang et~al.(2025)Yang, Li, Yang, Zhang, Hui, Zheng, Yu, Gao, Huang,
  Lv, et~al.]{yang2025qwen3}
An~Yang, Anfeng Li, Baosong Yang, Beichen Zhang, Binyuan Hui, Bo~Zheng, Bowen
  Yu, Chang Gao, Chengen Huang, Chenxu Lv, et~al.
\newblock {Qwen3 Technical Report}.
\newblock \emph{arXiv preprint arXiv:2505.09388}, 2025.
\newblock \doi{10.48550/arXiv.2505.09388}.
\newblock URL \url{https://arxiv.org/abs/2505.09388}.

\bibitem[Ye et~al.(2025)Ye, Xie, Zheng, Gao, Wu, Jiang, Li, and
  Kong]{ye2025dream}
Jiacheng Ye, Zhihui Xie, Lin Zheng, Jiahui Gao, Zirui Wu, Xin Jiang, Zhenguo
  Li, and Lingpeng Kong.
\newblock {Dream 7B: Diffusion Large Language Models}.
\newblock \emph{arXiv preprint arXiv:2508.15487}, 2025.
\newblock \doi{10.48550/arXiv.2508.15487}.
\newblock URL \url{https://arxiv.org/abs/2508.15487}.

\end{thebibliography}
\endgroup
\clearpage
\appendix
\setcounter{table}{0}
\setcounter{figure}{0}
\setcounter{equation}{0}
\renewcommand{\thetable}{A\arabic{table}}
\renewcommand{\thefigure}{A\arabic{figure}}
\renewcommand{\theequation}{A\arabic{equation}}
\section{Superimposed Sequence: Construction, Cost, and Discussion}
\label{app:sup}

\Needspace{5\baselineskip}
\subsection{Construction and Position Identifiers}

At decoding step $t$ the model input is the state $\mathbf{x}^{t-1}\in\mathcal{V}^{L}$
with masked index set $M_t$, i.e.\ the positions of the current
semi-autoregressive block that are still to be decoded. For every $i\in M_t$ the
pruned candidate set $\mathcal{S}^t_i$ of Eq.~(8) is already available from the
forward pass at step $t-1$; write $m_i=|\mathcal{S}^t_i|$ and
$\mathcal{S}^t_i=\{v_{i,1},\dots,v_{i,m_i}\}$.

Following Eq.~(9), the superimposed sequence $\mathbf{x}^t_{\mathrm{sup}}$
keeps $\mathbf{x}^{t-1}$ unchanged and then appends, for every $i\in M_t$ in
increasing order of $i$, the token $[x^{t-1}_i]$ --- a copy of the mask token,
written $\bar{i}$ --- followed by the candidate tokens
$(v_{i,1},\dots,v_{i,m_i})$. Let $c_{i,k}$ be the appended row containing
$v_{i,k}$ and $C_i=\{c_{i,1},\dots,c_{i,m_i}\}$ its candidate-row set.
We use zero-based row indices, matching Fig.~\ref{fig:attn_mask}. Three index
sets describe the construction:

\begin{itemize}
\itemsep2pt
\item the \emph{original sequence} $O=\{0,\dots,L-1\}$, carrying $\mathbf{x}^{t-1}$ unchanged;
\item the \emph{mask copies} $\bar{M}=\{\bar{i}: i\in M_t\}$, one per masked position;
\item the \emph{candidates} $C=\bigcup_{i\in M_t} C_i$.
\end{itemize}

The appended tokens of $\bar{M}$ and $C$ are interleaved position by position
and therefore do not form contiguous spans.

Position identifiers are assigned by
\begin{equation}
\begin{aligned}
\mathrm{pos}(o)&=o \;\;(o\in O),\\
\mathrm{pos}(\bar{i})&=\mathrm{pos}(c_{i,k})=i ,
\end{aligned}
\end{equation}
so that every appended token reuses the position identifier of the masked slot
it speaks for. The total length is
\begin{equation}
\label{eq:len}
L_{\mathrm{sup}} \;=\; L \;+\; \sum_{i\in M_t}\bigl(1+m_i\bigr).
\end{equation}
Because visibility is determined by the attention mask and locality by the
position identifiers, the physical ordering inside the appended region does not
affect the result.

The augmented input superposes candidate assignments at shared position IDs,
representing their collective influence in one forward pass. A position with
an empty candidate set still has a mask copy that probes candidates from other
positions, as illustrated by $D'$ in Figs.~\ref{overview} and~\ref{fig:attn_mask}.

\Needspace{5\baselineskip}
\subsection{Attention Mask}

Let $A\in\{0,1\}^{L_{\mathrm{sup}}\times L_{\mathrm{sup}}}$, where $A[a,b]=1$
means that query row $a$ may attend to key row $b$. The mask in
Fig.~\ref{fig:attn_mask} has three row types: original rows use full
bidirectional attention within $O$; candidate rows read $O$ and themselves;
mask-copy rows read $O$, themselves, and candidate rows associated with other
positions. Its complete visibility rule is
\begin{equation}
\label{eq:mask}
\begin{aligned}
A[a,b]&=\\
&\begin{cases}
1 & a\in O,\; b\in O,\\[2pt]
1 & a\in C,\; b\in O\cup\{a\},\\[2pt]
1 & \begin{gathered}a=\bar{i},\\
b\in O\cup\{\bar{i}\}\cup\!\bigcup_{j\ne i} C_j,\end{gathered}\\[2pt]
0 & \text{otherwise.}
\end{cases}
\end{aligned}
\end{equation}
For the six-position example in Fig.~\ref{fig:attn_mask},
$C_0=\{A_1,A_2\}$, $C_1=\{B_1\}$, and $C_3=\varnothing$.
Thus $A'$ reads $O\cup\{A',B_1\}$; $B'$ reads
$O\cup\{B',A_1,A_2\}$; and $D'$ reads
$O\cup\{D',A_1,A_2,B_1\}$. Each candidate reads $O$ and its own row.
Two properties of Eq.~\eqref{eq:mask} follow.

\paragraph{Path isolation.} Rows $a\in O$ have no support outside $O$, so the
original inference path is unaffected by anything that is appended.

\paragraph{Self-blinding.} Row $\bar{i}$ reads the candidates of other
positions while excluding $C_i$ and other mask-copy rows. Candidate rows
themselves read only $O$ and their own state. These two restrictions keep a
position's own candidates out of its probe's context across transformer
layers: foreign candidates obtain their contextual representations from $O$
and themselves, preserving the position-specific exclusion in the diagram.

Predictions are read at two disjoint sets of rows: the unperturbed distribution
$p_i$ at the original row $i\in O$, and the perturbed distribution
$\widetilde{p}_i$ at the mask-copy row $\bar{i}$. Candidate rows contribute
contextual keys and values to the probes. Their hidden states are updated
at every transformer layer under the second case of Eq.~\eqref{eq:mask};
their final vocabulary predictions are unused. This row rule is part of
the mask and determines the candidate representations consumed by the probes.

\Needspace{5\baselineskip}
\subsection{LEAP Contains Confidence-Based Decoding}

\begin{proposition}
\label{prop:exact}
Under Eq.~\eqref{eq:mask}, for every $a\in O$ the hidden states at every layer
and the output distribution produced by one forward pass on
$\mathbf{x}^t_{\mathrm{sup}}$ are identical to those produced by a forward pass
on $\mathbf{x}^{t-1}$ with the same full bidirectional attention within $O$.
\end{proposition}

\begin{proof}[Proof sketch]
By induction over layers. The attention softmax is normalised over unmasked
keys only, so keys with $A[a,b]=0$ contribute neither to the numerator nor to
the denominator. By the first case of Eq.~\eqref{eq:mask} together with its
default value $0$, layer-$\ell$ states at $O$ depend only on layer-$(\ell-1)$
states at $O$. The
position identifiers of $O$ are also unchanged, and the remaining sublayers act
position-wise.
\end{proof}

Two practical consequences follow. First, the quantity $p_i$ used in
Algorithm~1 is \emph{exactly} the prediction confidence-based parallel decoding
(CBPD) would compute, so the high-confidence set $\mathcal{H}_t$ recovered by
LEAP is exactly CBPD's; the lookahead machinery can only add positions, never
remove them. Writing $\mathcal{H}^{c}_t=\{i\in M_t: c_i\geq c\}$, with $\geq$ as in
Algorithm~1, and recalling
$\tau<\phi$, the decoded set of LEAP is sandwiched:
\begin{equation}
\label{eq:sandwich}
\mathcal{H}^{\phi}_t \;\subseteq\; \mathcal{U}_t \;\subseteq\; \mathcal{H}^{\tau}_t .
\end{equation}
Eq.~\eqref{eq:sandwich} holds when $\mathcal{H}^{\tau}_t\neq\varnothing$,
because even the most confident position $g_t$ then belongs to this set.
If $\mathcal{H}^{\tau}_t=\varnothing$, LEAP instead commits $g_t$, matching
CBPD's progress rule at the same state. These are per-step set relations on a
shared input, not a bound on total steps along different decoding trajectories.
Table~\ref{tab:tau} empirically shows more steps than CBPD at the same loose
threshold and fewer than CBPD at $\phi=0.9$.

Second, path isolation keeps the original context states separate from the
temporary lookahead states. The original path supplies the states used by
the backbone's context cache, while appended tokens are recomputed at each
step and stay outside that cache. This separation supports integration with
backbone caching optimizations, as illustrated by the LLaDA2.0-mini results
in the main paper.

\Needspace{5\baselineskip}
\subsection{Cost Analysis}

\paragraph{A hard bound on the candidate sets.} Since
$\sum_{v\in\mathcal{V}}p_\theta(x_i=v\mid\cdot)=1$ and every retained candidate
carries probability at least $\eta$,
\begin{equation}
m_i \;\leq\; \lfloor 1/\eta \rfloor ,
\end{equation}
independently of the vocabulary size. With the default $\eta=0.2$ this gives
$m_i\leq5$. Combined with $|M_t|\leq B$ for block size $B$, Eq.~\eqref{eq:len}
yields a worst case of $B(1+\lfloor 1/\eta\rfloor)=192$ appended tokens at
$B=32,\ \eta=0.2$ --- a constant that does not grow with the context length $L$.

\paragraph{Relative overhead.} Let
$\rho_t=\bigl(\sum_{i\in M_t}(1+m_i)\bigr)/L$. Attention cost grows as
$(1+\rho_t)^2$ and the position-wise sublayers as $(1+\rho_t)$. In practice
$\rho_t$ stays far below the worst case, for two reasons: most masked positions
have a peaked predictive distribution and contribute one or zero candidates, and
$|M_t|$ shrinks monotonically as the block is decoded.

\paragraph{Measured overhead.} Table~\ref{tab:overhead} reports the per-step
cost on LLaDA-8B-Instruct at the two extremes of the $\eta$ range; the default
$\eta=0.2$ lies between them. Sequence length grows by at most $3.6\%$ and
per-step latency by at most $9.0\%$. Since LEAP removes roughly $30\%$ of the
denoising steps, the product is a net reduction of $20\%$--$30\%$ in total token
forward operations, consistent with the $\eta$ sweep in
Table~\ref{tab:eta}.

\begin{center}
\begin{minipage}{\textwidth}
\captionsetup{hypcap=false}
\centering
\small
\setlength{\tabcolsep}{3.3pt}
\begin{tabular}{llcccc}
\toprule
Dataset & Setting & Len. & $\Delta$ & Lat./step & $\Delta$ \\
\midrule
\multirow{3}{*}{HumanEval}
 & Vanilla    & 850  & --      & 91.1\,ms  & --      \\
 & $\eta=0.5$ & 860  & $+1.2\%$ & 94.9\,ms  & $+4.2\%$ \\
 & $\eta=0.1$ & 877  & $+3.2\%$ & 99.3\,ms  & $+9.0\%$ \\
\midrule
\multirow{3}{*}{GSM8K}
 & Vanilla    & 1135 & --      & 108.6\,ms & --      \\
 & $\eta=0.5$ & 1151 & $+1.4\%$ & 114.2\,ms & $+5.2\%$ \\
 & $\eta=0.1$ & 1176 & $+3.6\%$ & 118.3\,ms & $+8.9\%$ \\
\bottomrule
\end{tabular}
\captionof{table}{Per-step sequence length and latency of the superimposed forward pass
on LLaDA-8B-Instruct, averaged over the decoding trajectory. Percentages are
relative to the vanilla forward pass on the same sequence.}
\label{tab:overhead}
\end{minipage}
\end{center}

\section{Complete Hyperparameter Sweeps}
\label{app:sweeps}

\begin{figure*}[!htbp]
\centering
\includegraphics[width=0.85\textwidth]{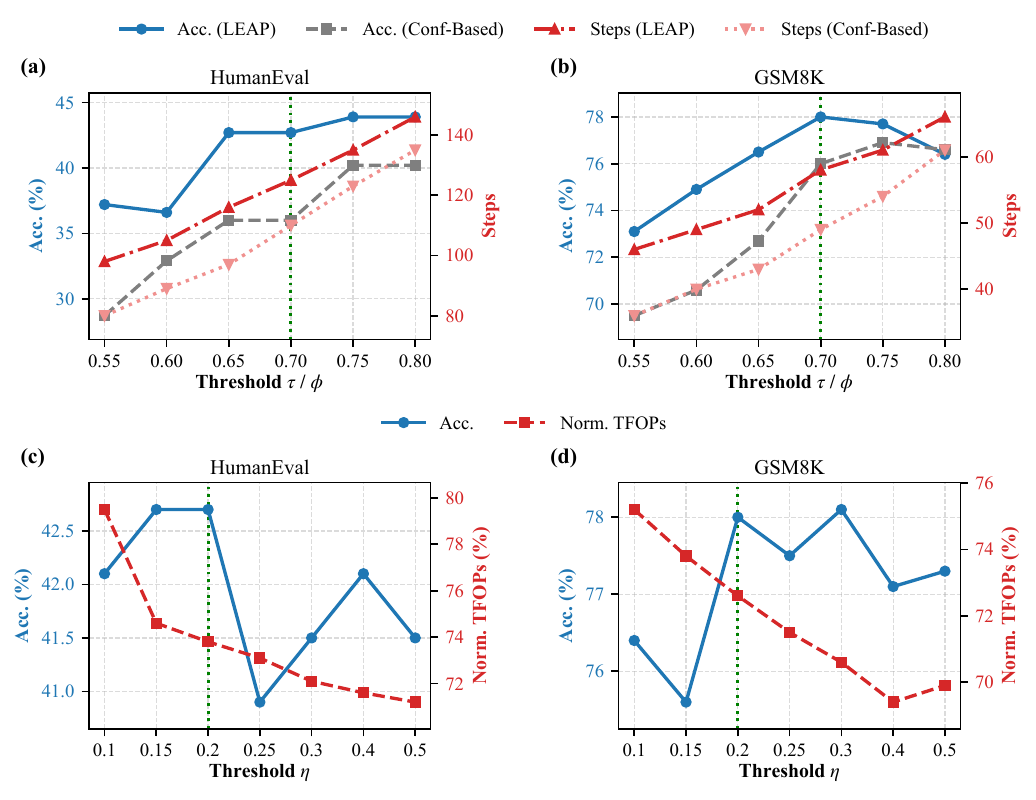}
\caption{Complete hyperparameter sweeps on LLaDA-8B-Instruct. (a, b) accuracy
and denoising steps versus the decoding threshold, for LEAP (threshold used as
$\tau$) and confidence-based decoding (threshold used as $\phi$). (c, d)
accuracy and normalised TFOPs versus the pruning threshold $\eta$. Dotted green
lines mark the defaults $\tau=0.7$ and $\eta=0.2$. The $\eta$ axis is
categorical because the grid is non-uniform.}
\label{fig:sweeps}
\end{figure*}

The main paper reports the sensitivity study on HumanEval only. Here we give
both benchmarks, both thresholds, and --- for $\tau$ --- confidence-based
decoding evaluated at the \emph{same} threshold value, which isolates the
contribution of the consistency test from that of the looser threshold. All
runs use LLaDA-8B-Instruct with block size $32$, greedy sampling and batch
size $1$; when one threshold is swept the other is held at its default
($\eta=0.2$, $\tau=0.7$). Figure~\ref{fig:sweeps} plots the same data.

\Needspace{5\baselineskip}
\subsection{Effect of the Decoding Threshold \texorpdfstring{$\tau$}{tau}}

\begin{center}
\begin{minipage}{\textwidth}
\captionsetup{hypcap=false}
\centering
\small
\setlength{\tabcolsep}{3.4pt}
\begin{tabular}{lcccccc}
\toprule
& \multicolumn{3}{c}{HumanEval} & \multicolumn{3}{c}{GSM8K} \\
\cmidrule(lr){2-4}\cmidrule(lr){5-7}
Thr. & LEAP & Conf & $\Delta$ & LEAP & Conf & $\Delta$ \\
\midrule
0.55 & 37.2 & 28.7 & $+8.5$ & 73.1 & 69.5 & $+3.6$ \\
0.60 & 36.6 & 32.9 & $+3.7$ & 74.9 & 70.6 & $+4.3$ \\
0.65 & 42.7 & 36.0 & $+6.7$ & 76.5 & 72.7 & $+3.8$ \\
0.70 & 42.7 & 36.0 & $+6.7$ & 78.0 & 76.0 & $+2.0$ \\
0.75 & 43.9 & 40.2 & $+3.7$ & 77.7 & 76.9 & $+0.8$ \\
0.80 & 43.9 & 40.2 & $+3.7$ & 76.4 & 76.6 & $-0.2$ \\
\midrule
\multicolumn{7}{l}{\emph{Denoising steps at the same threshold}} \\
0.55 & 98  & 80  & $+18$ & 46 & 36 & $+10$ \\
0.60 & 105 & 89  & $+16$ & 49 & 40 & $+9$  \\
0.65 & 116 & 97  & $+19$ & 52 & 43 & $+9$  \\
0.70 & 125 & 110 & $+15$ & 58 & 49 & $+9$  \\
0.75 & 135 & 123 & $+12$ & 61 & 54 & $+7$  \\
0.80 & 146 & 135 & $+11$ & 66 & 61 & $+5$  \\
\bottomrule
\end{tabular}
\captionof{table}{Accuracy (\%) and denoising steps as a function of the threshold, on
LLaDA-8B-Instruct. ``LEAP'' uses the value as $\tau$; ``Conf'' is
confidence-based decoding using the same value as $\phi$. $\Delta$ is
LEAP\,$-$\,Conf. Step counts are averages over the evaluation set. For
reference, confidence-based decoding at its default $\phi=0.9$ reaches $42.1$ in
$176$ steps on HumanEval and $76.5$ in $79$ steps on GSM8K.}
\label{tab:tau}
\end{minipage}
\end{center}

\paragraph{The consistency test, not the looser threshold, carries the gain.}
At every threshold except one, LEAP is more accurate than confidence-based
decoding operating at the identical value, by up to $8.5$ points on HumanEval
and $4.3$ points on GSM8K. At a shared input state, both draw threshold-driven commitments from
$\mathcal{H}^{\tau}_t$. LEAP retains high-confidence positions and adds those
passing the lookahead test, together with the shared most-confident progress
rule. Their subsequent contexts can differ.

\paragraph{The advantage is largest in the medium-confidence band.} As the
threshold approaches $\phi$, $\mathcal{H}^{\tau}_t$ shrinks towards
$\mathcal{H}^{\phi}_t$ and the sandwich of Eq.~\eqref{eq:sandwich} tightens; the
gap accordingly narrows from $+3.6$ to $-0.2$ on GSM8K between $0.55$ and
$0.80$. 

\paragraph{LEAP reaches a point that thresholding cannot.} On GSM8K, LEAP at
$\tau=0.7$ attains $78.0$ in $58$ steps, whereas confidence-based decoding at
$\phi=0.9$ attains $76.5$ in $79$ steps: LEAP is simultaneously faster and more
accurate. The same holds on HumanEval ($42.7$ in $125$ steps versus $42.1$ in
$176$ steps). Lowering $\phi$ does not recover this operating point, since
confidence-based decoding at $0.7$ collapses to $36.0$ on HumanEval.

The two benchmarks differ in shape. HumanEval degrades sharply below $\tau=0.65$
and then saturates, whereas GSM8K is unimodal with a maximum at $\tau=0.70$ and
a mild decline afterwards. We therefore adopt $\tau=0.7$, which is the peak on
GSM8K and sits at the start of the plateau on HumanEval; pushing to $\tau=0.8$
buys $1.2$ points on HumanEval at the cost of $17\%$ more steps and loses $1.6$
points on GSM8K.

\Needspace{5\baselineskip}
\subsection{Effect of the Pruning Threshold \texorpdfstring{$\eta$}{eta}}

\begin{center}
\begin{minipage}{\textwidth}
\captionsetup{hypcap=false}
\centering
\small
\setlength{\tabcolsep}{4.2pt}
\begin{tabular}{lcccc}
\toprule
& \multicolumn{2}{c}{HumanEval} & \multicolumn{2}{c}{GSM8K} \\
\cmidrule(lr){2-3}\cmidrule(lr){4-5}
$\eta$ & Acc. & TFOPs & Acc. & TFOPs \\
\midrule
0.10 & 42.1 & 79.5 & 76.4 & 75.2 \\
0.15 & \textbf{42.7} & 74.6 & 75.6 & 73.8 \\
0.20 & \textbf{42.7} & 73.8 & 78.0 & 72.6 \\
0.25 & 40.9 & 73.1 & 77.5 & 71.5 \\
0.30 & 41.5 & 72.1 & \textbf{78.1} & 70.6 \\
0.40 & 42.1 & 71.6 & 77.1 & 69.4 \\
0.50 & 41.5 & 71.2 & 77.3 & 69.9 \\
\bottomrule
\end{tabular}
\captionof{table}{Accuracy (\%) and total token forward operations as a function of the
pruning threshold $\eta$, on LLaDA-8B-Instruct. TFOPs are normalised to
confidence-based decoding at $\phi=0.9$ ($=100$), so every entry below $100$
denotes a net computational saving. Bold marks the best accuracy per benchmark.}
\label{tab:eta}
\end{minipage}
\end{center}

Table~\ref{tab:eta} shows that accuracy is comparatively insensitive to $\eta$:
the spread is $1.8$ points on HumanEval and $2.5$ on GSM8K, against $6.7$ and
$2.0$ points for the consistency test itself at the same threshold. Cost
decreases almost monotonically with $\eta$, since a larger $\eta$ admits fewer
candidates per position and therefore a shorter superimposed sequence.

Two further properties bear on the choice of default. First, every entry lies between $69$ and $80$: even
at the most permissive setting $\eta=0.1$, where the superimposed sequence is
longest, LEAP spends $20\%$--$30\%$ fewer token forward operations than
confidence-based decoding, because the reduction in denoising steps dominates
the per-step overhead of Table~\ref{tab:overhead}. Second, $\eta=0.2$ is the
best joint operating point. It attains the (tied) best accuracy on
HumanEval\,---\,$42.7$, shared with $\eta=0.15$ but at lower cost\,---\,and
comes within $0.1$ points of the best on GSM8K, where $\eta=0.30$ leads with
$78.1$ against $78.0$. Buying that $0.1$ would give up $1.2$ points on
HumanEval, so we use $\eta=0.2$ as the default throughout the main paper.

\end{document}